\RequirePackage{fix-cm}
\documentclass[smallextended]{svjour3}       
\usepackage[utf8]{inputenc}
\usepackage{tikz-dependency}
\usepackage{float}
\usepackage{tipa}
\usepackage{multirow}
\usepackage{supertabular,array,booktabs}
\usepackage{relsize}
\usepackage{lingmacros}
\usepackage{fnpct}
\usepackage{cjhebrew}

\usepackage{xcolor}
\definecolor{darkblue}{rgb}{0, 0, 0.5}
\usepackage{color}
\usepackage{bm}
\definecolor{orange}{rgb}{1,0.5,0}
\definecolor{mdgreen}{rgb}{0,0.6,0}
\definecolor{mdblue}{rgb}{0,0,0.7}
\definecolor{dkblue}{rgb}{0,0,0.5}
\definecolor{dkgray}{rgb}{0.3,0.3,0.3}
\definecolor{slate}{rgb}{0.25,0.25,0.4}
\definecolor{gray}{rgb}{0.5,0.5,0.5}
\definecolor{ltgray}{rgb}{0.7,0.7,0.7}
\definecolor{ltltgray}{rgb}{0.9,0.9,0.9}
\definecolor{purple}{rgb}{0.7,0,1.0}
\definecolor{lavender}{rgb}{0.65,0.55,1.0}

\usepackage{hyperref}
\hypersetup{colorlinks=true,citecolor=darkblue, linkcolor=darkblue, urlcolor=darkblue}

\usepackage{listings}
\usepackage{graphicx}
\setkeys{Gin}{width=\linewidth,totalheight=\textheight,keepaspectratio}
\graphicspath{{graphics/}}
\usepackage{caption}
\usepackage{subcaption}

\usepackage{url}

\usepackage[normalem]{ulem}
\usepackage{amsmath}
\usepackage{multirow}
\usepackage{multicol}
\usepackage{wrapfig}

\usepackage{tree-dvips}
\usepackage{qtree}
\usepackage{tikz}
\usetikzlibrary{shapes,arrows,fit,calc,er,positioning,intersections,decorations.shapes}

\usepackage{hhline}
\usepackage{colortbl}
\usepackage{appendix}
\usepackage{supertabular,array}

\tikzset{decorate sep/.style 2 args={decorate,decoration={shape backgrounds,shape=circle,shape size=#1,shape sep=#2}}}

\usepackage{nameref}
\usepackage{natbib}

\newcommand{\oamod}[1]{{{#1}}}
\newcommand{\msmod}[1]{{{#1}}}
\newcommand{\aumod}[1]{{{#1}}}
\newcommand{\nssmod}[1]{{{#1}}}

\usepackage[prependcaption,textsize=scriptsize]{todonotes}
\makeatletter 
\@mparswitchfalse%
\makeatother
\normalmarginpar 
\newcommand{\aunote}[1]{\todo[backgroundcolor=white,author=AuH2O]{#1}}
\newcommand{\nssnote}[1]{\todo[backgroundcolor=white,author=NSS]{#1}}
\newcommand{\oanote}[1]{\todo[backgroundcolor=white,author=OA]{#1}}
\renewcommand{\aunote}[1]{\unskip}
\renewcommand{\nssnote}[1]{\unskip}
\renewcommand{\oanote}[1]{\unskip}

\newcommand{\com}[1]{}

\newcommand{\lamwh}[2]{$\lambda$#1. #2}
\newcommand{\lamconj}[6]{$\lambda$\var{#2}. #1(#3\textsubscript{\var{#2}}(#4), #5\textsubscript{\var{#2}}#6)}
\newcommand{\lamand}[5]{\lamconj{and}{#1}{#2}{#3}{#4}{#5}}
\newcommand{\lamnest}[4]{$\lambda$\var{#1}. #2\textsubscript{\var{#1}}(#3\textsubscript{\var{#1}}(#4))}
\newcommand{\lam}[3]{$\lambda$\var{#1}. #2\textsubscript{\var{#1}}(#3)}
\newcommand{\lamn}[2]{$\lambda$#1. #2}
\newcommand{\npred}[5]{$\lambda$\var{#1} \quant{#2}{#3}{#4\textsubscript{\var{#1}}(#5)} }
\newcommand{\quanthelp}[3]{\textsc{#1} #2\big[#3\big]}
\newcommand{\quant}[3]{\quanthelp{\textsc{#1}}{#2}{#3}}
\newcommand{\var}[1]{e\textsubscript{#1}}
\newcommand{\lf}[1]{\emph{#1}}

\begin{document}

\title{Cross-linguistically Consistent Semantic and Syntactic Annotation of Child-directed Speech}


\titlerunning{Semantic and Syntactic Annotation of Child Directed
Speech}        

\author{
Ida Szubert
\and
%
Omri Abend
\and \\
Nathan Schneider
\and
Samuel Gibbon
\and
Louis Mahon
\and
Sharon Goldwater
\and
Mark Steedman
}

\authorrunning{Szubert et al.} 

\institute{Szubert, Mahon, Goldwater and Steedman are from the School of Informatics at the University of Edinburgh, UK; 
Gibbon is from the Centre for Clinical Brain Sciences at the University of Edinburgh, UK;
Abend is from the School of Computer Science and Engineering and the Department of Cognitive Science of the Hebrew University of Jerusalem, Israel; Schneider is from the Departments of Linguistics and Computer Science of Georgetown University, D.C., USA.\\
ORCID identifiers are Abend (0000-0003-4311-3876), Schneider (0000-0002-5994-671X), Gibbon (0000-0002-5485-7523), Mahon (0000-0003-0571-4611), Goldwater (0000-0002-7298-0947), Steedman (0000-0003-2509-0797).\\
Email for Correspondence:  {omri.abend@mail.huji.ac.il} (Abend)
}

\date{Received: date / Accepted: date}

\maketitle

\begin{abstract}

Corpora of child speech and child-directed speech (CDS) have enabled major contributions to the study of child language acquisition, yet semantic annotation for such corpora is still scarce and lacks a uniform standard. Semantic annotation of CDS is particularly important for understanding the nature of the input children receive and developing computational models of child language acquisition. For example, under the assumption that children are able to infer meaning representations for (at least some of) the utterances they hear, the acquisition task is to learn a grammar that can map novel adult utterances onto their corresponding meaning representations, in the face of noise and distraction by other contextually possible meanings. To study this problem and to develop computational models of it, we need corpora that provide both adult utterances and their meaning representations, ideally using annotation that is consistent across a range of languages in order to facilitate cross-linguistic comparative studies.

This paper proposes a methodology for constructing such corpora of CDS paired with sentential logical forms, and uses this method to create two such corpora, in English and Hebrew. 
The approach enforces a cross-linguistically consistent representation, building on recent advances in dependency representation and semantic parsing. 
Specifically, the approach involves two steps. First, we annotate the corpora using the Universal Dependencies (UD) scheme for syntactic annotation, which has been developed to apply consistently to a wide variety of domains and typologically diverse languages.
Next,
we further annotate these data by applying an automatic method for transducing sentential logical forms (LFs) from UD structures. The UD and LF representations have complementary strengths: 
UD structures are language-neutral and support consistent and reliable annotation by multiple annotators, whereas LFs are neutral as to their syntactic derivation and transparently encode semantic relations. 

Using this approach, we provide syntactic and semantic annotation for two corpora from CHILDES: Brown's Adam corpus (English; we annotate $\approx$80\% of its child-directed utterances), all child-directed utterances from Berman's Hagar corpus (Hebrew).
We verify the quality of the UD annotation using an inter-annotator agreement study, and manually evaluate the transduced meaning representations.
We then demonstrate the utility of the compiled corpora through (1) a longitudinal corpus study of the prevalence of different syntactic and semantic phenomena in the CDS, and (2) applying an existing computational model of language acquisition to the two corpora and briefly comparing the results across languages.


\end{abstract}


\section{Introduction}

As research in child language acquisition becomes increasingly data-driven, the availability of annotated corpora of child and child-directed speech (CDS) is increasingly important as a basis for understanding the process of child language acquisition from such input. The CHILDES project \citep{MacW:00} has been pivotal in the effort to
streamline data collection and to standardize linguistic annotation in
this domain. However, despite these achievements, CDS resources
annotated with semantic annotation are scarce, and lack a uniform standard.  
Indeed, even syntactic annotation is only available in CHILDES for a
handful of languages, and these are not all annotated according to the
same scheme. For example, \citet{Saga:10} developed a dependency annotation for
CHILDES and applied it to English and Spanish, whereas
\citet{Gretz:15} used a different dependency scheme when annotating
Hebrew CDS. 
Neither of these schemes is standardly used in the field of natural language processing (NLP), limiting
the application of NLP tools developed elsewhere. 
\aumod{Meanwhile, the Dutch AnnCor CHILDES Treebank \citep{odijk2018anncor} uses yet another dependency scheme, based on the Alpino parser \citep{bouma2001alpino}, and a sizeable portion of the English CHILDES Treebank has been annotated with constituency trees following the Penn Treebank annotation scheme \citep{pearl2013syntactic}.} 
\nssmod{Thus, when it comes to acquisition corpora, syntactic annotations are heterogeneous within and between languages, and do not necessarily reflect prevailing approaches for annotating other genres.}

\nssmod{Despite a number of linguistic challenges in analyzing transcribed speech of adult-child interactions, we argue that datasets for studying syntactic acquisition need not be idiosyncratic. This work investigates, first, whether a syntactic framework that is now well-established in NLP -- Universal Dependencies -- can be applied to child-directed speech transcripts in multiple languages; and second, how language-agnostic rules can map such annotations into sentential logical forms suitable for studying the Semantic Bootstrapping Hypothesis, and the acquisition of grounded sentential semantics \citep[e.g.,][]{Mao:21a}.

We motivate the components of this goal in turn.}

\paragraph{Child-directed speech.} In approaching syntax and semantics in acquisition data, we are mindful of the fact that empirical studies of language acquisition often focus on child-directed speech, i.e.,
utterances by adults who are interacting with the child
learner. \msmod{Despite the fact that the child's own utterances are in fact annotated in the original CHILDES corpora, we follow the above research in further annotating only the child-directed side of the
data, leaving the child's own utterance unaffected, for two reasons.}
The first is that it is the child-directed component of the dialog that provides the language-specific input to the child's language-learning process, and the data for any model of how that process works. The second is that almost the only thing that we know about the structures or meaning representations that underlie early
child utterances is that they are continuously changing -- and thus, in
our view, best modeled as latent structure.



\paragraph{\nssmod{Cross-linguistic applicability.}}
To the best of our knowledge, the present work is the first to
apply a cross-linguistically consistent syntactic \aumod{annotation scheme to CDS. This consistency is important to enable comparisons across typologically distinct languages: both corpus analyses investigating features of the adult input, and modelling studies testing theories of language acquisition.}
To illustrate its use, we annotate corpora in two languages: English and Hebrew. 
We also propose a methodology for producing cross-linguistically consistent {\it semantic} annotation of CDS. 

\paragraph{\nssmod{Syntactic framework.}}
As a syntactic representation, from which we will generate the non-aligned logical forms that provide the input to the child or computational learning model, we use the Universal Dependencies (UD) standard
\citep{nivre2016universal,de_marneffe-21}, motivated by its demonstrated
applicability to a wide variety of domains and languages, and its
relative  reliability for manual annotation of
corpora \citep{berzak2016anchoring}. Moreover, as UD is the de facto standard for dependency annotation in NLP, it is supported by a
large and expanding body of research work, and by a variety of parsers and other tools. The UD standard is briefly presented in \S\ref{sec:UD}. 
\nssmod{Our annotation reveals various distinctive characteristics of the CDS genre, for which we propose UD conventions (\S\ref{sec:ud_in_cds}).}

\paragraph{\nssmod{Logical forms and semantic bootstrapping.}}
Sentential logical forms (henceforth, LFs) are an essential building block in a
complete linguistic analysis of CDS, and are needed for computational
implementations of theories of acquisition that emphasize the role of ``semantic bootstrapping'', i.e., theories that construe grammar acquisition as the attachment of language-specific syntax to logical forms related to a
universal conceptual structure \cite[e.g.,][]{Bowerman:74, Culi:84, Pink:79,Bris:00,Buttery:06,Abend:17}.
Nevertheless, very few corpora of CDS are annotated with sentential meaning representations. 
Examples include verb- and preposition-sense annotation, as well as
semantic role-labeling of data from English CHILDES by \citet{moon2018gold}, 
and sentential logical forms produced by \citet{Vill:02}, \citet{Buttery:06} and by \citet{Kwia:12}.
A related line of work automatically generated inputs for computational models of acquisition from a semantic lexicon \citep{alishahi2008computational}.
We are not aware of any semantically annotated CDS corpora for
languages other than English. 
To address this gap, we further propose a method for automatically transducing
LFs from UD structures, thereby obtaining cross-linguistic consistency
for those annotations as well, while 
avoiding the difficult and error-prone procedure of annotating LFs
over utterances from scratch. 

\paragraph{\nssmod{Semantics beyond syntax.}}
Although the LF level of representation is deterministically derived from the dependency level, this additional level of annotation is important since it is neutral with respect to surface word order and therefore comparatively language-independent -- a key feature for developing and testing models of language acquisition.
The transduction process we propose therefore abstracts away from
syntactic detail, and transparently encodes information which
is implicit in UD -- in particular, long-range dependencies. 
As an example, consider the following, in which the
subject ``you'' and the object ``it'' are shared between ``find'' and 
``bring'':\footnote{\label{fn:words-constants}\nssmod{For simplicity, we notate most conceptual content in the LF as words (e.g., \textit{you} rather than $\textit{you}'$), to be understood as logical constants.}}

\begin{center}
\begin{dependency}
 	\begin{deptext}
	 you \& find \& and \& bring \& it \\
 	\end{deptext}
 \deproot[edge unit distance=2.3ex]{2}{root}
 \depedge{2}{1}{nsubj}
 \depedge{4}{3}{cc}
 \depedge{2}{4}{conj}
 \depedge{4}{5}{dobj}
 \end{dependency}
\end{center}
\enumsentence{
\centering
LF: \lf{ \lamand{1}{find}{you, it}{bring}{(you, it)} }
\label{you_bring_it}
}
%
This information  is only implicit in the UD structure, but is made
explicit in the LF (though see \S\ref{sec:limitations}). As is the case
with any dependency annotation, some distinctions (such as coordination
and scope) are underspecified in UD. We disambiguate some of these
cases by refining the set of UD labels (see \S\ref{sec:UD}). \oamod{Other cases cannot be handled effectively due to their underspecification in the UD formalism (as opposed to other grammar formalisms, such as, e.g., CCG \citep{stee:99}, or semantic schemes such as AMR \citep{Banarescu:12}. We discuss the relationship between our LF formalism and other semantic schemes in \S\ref{sec:ud-lf} and discuss its limitations in \S\ref{sec:limitations}.}

The conversion method is implemented by recursively building the LFs
using unlexicalized rules that condition only on the UD dependency
tree and Part of Speech (POS) tags.\footnote{The only exceptions are the wh-pronouns,
  which are lexically conditioned.} 
As such, these rules can be applied to any UD-annotated sentence,
regardless of its language.  
In this we follow the framework of \citet{reddy2016transforming}, but
cover a wider range of semantic phenomena, using a different
representation language.\footnote
{Reddy's representation is specialized for querying the Google/FreeBase Knowledge Graph.}

\paragraph{Nature of the LFs.} Our LFs, detailed in \S\ref{sec:ud-lf}, reflect what we take to be a fairly standard model-theoretic semantics.  
The focus is on compositional, as opposed to lexical, aspects of sentence meaning -- i.e., aspects most crucial to modeling the acquisition of syntax.
\nssmod{Notably, in NLP there is a wider landscape of symbolic} meaning representations \nssmod{applied to corpora,} such as Universal Conceptual Cognitive Annotation \citep[UCCA;][]{abend2013universal}, Abstract Meaning Representations \citep[AMR;][]{amr}, and the Generative Lexicon \citep[GL;][]{pustejovsky1998generative}. 
\nssmod{Those representations, however, contain additional elements of meaning (like coreference and richer lexical semantics), and are therefore more challenging to annotate or parse.}\footnote{\nssmod{AMR, moreover, was initially designed just for English, and without anchoring of concepts to words in the sentence, which makes it challenging to derive an AMR graph compositionally \citep[attempts to do so include][]{szubert-etal-2018-structured,groschwitz-18,blodgett-19,leamr}. A new framework, Uniform Meaning Representation \citep[UMR;][]{umr}, aims to address some of these limitations, but is still under development.}}
\nssmod{Our LFs could, however, provide a starting point for inducing more elaborate semantic annotations in such frameworks.}

\paragraph{\nssmod{New resource.}}
Using the proposed protocol of syntactic annotation, we annotate a
large contiguous portion of Brown's Adam corpus from CHILDES (the
first $\approx$80\% of its child-directed utterances, comprising over 17K English utterances), as well as over 24K Hebrew utterances, constituting the
entire Hagar CHILDES corpus
\citep{berman1990acquiring}. 
The corpora were selected for their sizes, which are large for CDS
corpora, and because they have an initial (non-UD) dependency annotation, part manual and part automatic, which makes our UD annotation process easier \citep{Saga:10} (see below).
In addition, the Adam Corpus was chosen because of the
availability of the other labeled versions
\citep{pearl2013syntactic,moon2018gold}, and because of the large
amount of psycholinguistic study that has been applied to it
(\citealp{McNeill:66,Brow:73}, {\em passim}).  

To obtain gold-standard UD trees, we take advantage of the existing
syntactic annotations in these corpora: 
we automatically convert them into approximate UD trees (\S\ref{sec:procedure}), 
then hand-correct the converted outputs.
We chose this procedure as we found it to be much faster than annotation from scratch, but note that it is not required: other corpora without preexisting dependency annotation could be annotated with UD parses directly.
A schematic overview of the complete syntactic/semantic annotation methodology is given in Figure~\ref{fig:schema}. 

We note that \citet{liu-21}, in contemporaneous work, annotated the English Eve corpus with UD structures, using a semi-automatic approach akin to ours (but did not address other languages or the transduction of logical forms).

\begin{figure}
\footnotesize
\centering
\tikzstyle{decision} = [diamond, draw, fill=blue!20, 
    text width=4.5em, text badly centered, node distance=3cm, inner sep=0pt]
\tikzstyle{block} = [rectangle, draw, fill=blue!20, 
    text width=5em, text centered, rounded corners, minimum height=4em]
\tikzstyle{line} = [draw, -latex']
\tikzstyle{cloud} = [draw, ellipse,fill=red!20, node distance=3cm,
    minimum height=2em]
    
\begin{tikzpicture}[node distance = 2cm, auto]
    \node [block] at (0,0) (init) {Transcribed CDS};
    \node [block] at (-2,-3) (scratch) {Annotate UD from scratch};
    \node [block] at (2,-2) (convert) {Convert to UD};
    \node [block] at (2,-4) (correct) {Hand-correct UD parses};
    \node [block] at (0,-6) (LF) {Transduce LFs from UD parses};

  	\draw[->] (init) -- (scratch) node[above,yshift=1.5cm,xshift=.2cm] {if not parsed};
    \draw[->] (init) -- (convert) node[above,yshift=.8cm] {if parsed};
    \path[line] (convert) -- (correct);
    \path[line] (correct) -- (LF);
    \path[line] (scratch) -- (LF);
\end{tikzpicture}
\caption{Main stages of the proposed annotation methodology.
\label{fig:schema}}
\end{figure}

\paragraph{\nssmod{Evaluation.}}
We evaluate our method by first measuring inter-annotator agreement for UD parses in both corpora, showing that UD
can be reliably applied to CDS in both languages (\S\ref{sec:experiments}). Of all parsed sentences, our LF conversion tool is able to produce an output for 80.5\% (English) and 72.7\% (Hebrew). We then manually evaluate a small sample of these LFs and find that 82\% of the LFs in both languages are fully correct. Most errors fall into a small number of categories, discussed in \S\ref{sec:limitations}. 

Next, we provide some simple proof-of-concept analyses illustrating the benefit of these cross-linguistically consistent annotations (\S\ref{sec:analyses}). We compare the usage frequency of different dependency types in our CDS corpora relative to written text corpora in the same languages, and between the English and Hebrew CDS corpora. Overall, we find that the CDS corpora are more similar to each other than to the text corpora in the same language. We also perform a longitudinal analysis, looking for systematic changes in the frequency of use of various syntactic constructions. We find that while in the English corpus only a small number of constructions increase in frequency (adjectival and relative clauses, noun compounding, and noun ellipsis), in the Hebrew one the changes are much more widespread. This can possibly be explained by the different ages of the children at the time of data collection. \oamod{The finding for English could be relevant to the ongoing discussion as for whether the complexity of CDS changes or not over the longitudanal trajectory. Our findings can be interpreted as echoing the findings of \citet{Newp:77}, who also found that syntactic complexity in English CDS does not generally increase with time, except for the number of clauses, which shows a moderate increase.} 

Finally, \oamod{as a proof of concept for demonstrating the utility of this work for the modeling of child language acquisition, 
we adapt the acquisition learning model by \citet{Abend:17} to learn from the transduced LFs (\S\ref{sec:simulations}). Experiments are conducted for both English and Hebrew. Results show qualitatively similar trends to the ones reported by \citet{Abend:17}.}

\vspace{.25\baselineskip}
To recap, we present the following contributions:
\begin{enumerate}
\item
    We show that the UD scheme can be applied to CDS \nssmod{with some additional guidelines}, and conduct an inter-annotator agreement study to confirm this finding. 
\item
    We compile two UD-annotated corpora of CDS, one in English and one in Hebrew.
\item
    We develop an automatic conversion method and codebase for converting UD-annotated CDS to logical forms.
\item
    We perform a longitudinal corpus study of the prevalence of different syntactic and semantic phenomena in CDS, across the two languages.

\item
    We show that a baseline grammar for both languages can be induced from the CDU-LF pairs in the corpora by the learner of \cite{Abend:17}.
\end{enumerate}

Our \nssmod{annotated data and transduction code} are available at \url{https://github.com/ida-szubert/CHILDES_UD2LF}. The code for running the simulations is available at \url{https://github.com/ida-szubert/ccg_acquisition_2}.\footnote{The repositories are still updated from time to time, with small improvements/revisions (and the commit history is of course retained for reproducibility).}

\section{The Universal Dependencies Scheme}\label{sec:UD}

Universal Dependencies \cite[UD]{nivre2016universal,de_marneffe-21} is a coarse-grained
syntactic dependency scheme which has quickly become the de facto standard
for annotating dependencies in many languages.
It is designed to establish a unified standard for dependency annotation 
across languages and domains,
to support rapid annotation, and to be suitable for parsing and helpful 
for downstream language understanding tasks. 
All these design principles fit naturally with the goals of this paper.
Moreover, in order to attain cross-linguistic applicability, UD's design conventions are often similar to those made by semantic schemes \citep{hershcovich-etal-2019-content}.

Formally, UD uses trees in which nodes are lexical items and directed edges represent dependencies labeled with types such as subject, modifier, etc. UD further includes conventions for annotating morphology, although only POS tags, morphological features and dependency structures are addressed in this work.\footnote{\nssmod{Our rules do not invoke specific morphological features, but we retain morphological annotations from CHILDES: see \S\ref{sec:conversion}.}} We use the UD guidelines version 1.0, as reference corpora for version 2.0 were not available at the time of annotation.\footnote{The only exception to this rule is that we attach coordinating conjunctions (\textit{cc}) to their following conjunct, as in version 2.0.
Note that it is straightforward to convert this convention to the version 1.0 convention (attach each \textit{cc} edge to the parent of its endpoint), where doing so inversely is non-trivial.} 

We will now turn to UD's treatment of frequent constructions. 
A glossary of some common UD edge types used in this paper is given in Table \ref{tab:categories}.

Throughout the rest of the paper we will use the CHILDES transliteration scheme for Hebrew, which directly reflects the writing system of Hebrew.

\begin{table}
\centering
\small
\begin {tabular}{|c|p{10cm}|}
\hline
Label & Short Definition\\
\hline
\hline
\multicolumn{2}{|c|}{{\bf Clause Elements}}\\
\hline
 {\it nsubj} & Nominal subject.\\
\hline
 {\it dobj} & Direct object.\\
\hline
 {\it ccomp} & Clausal complement (finite or infinite), unless its subject is controlled.\\
\hline
 {\it xcomp} & Open clausal complement, i.e., predicative or clausal complement without its own subject.\\
 \hline
 {\it advmod} & Modifying adverb.\\
 \hline
 {\it neg} & Negation modifier (e.g., ``not'', ``no'').\\
\hline
 {\it aux} & Auxiliary of a verbal predicate, including markers of tense, mood, modality, aspect, voice or evidentiality.\\
\hline
 {\it nmod} & Oblique: nominal functioning as an adjunct. ({\it nmod}s are also used for nominal modifiers in noun phrases, see below)\\
\hline \hline

\multicolumn{2}{|c|}{{\bf Inter-clause Linkage}}\\
\hline
 {\it conj} & Relation between the conjuncts in a coordination to the first conjunct, which is considered the head.\\
\hline
 {\it cc} & Coordinating conjunction.\\
\hline
 {\it advcl} & Adverbial clause modifier, including temporal clause, consequence, conditional clause, and purpose clause.\\
\hline
 {\it mark} & Marker: the word introducing a clause subordinate to another clause, often a subordinating conjunction. \\
\hline
 {\it parataxis} & Several elements (often clauses or fragments) placed side by side without any explicit coordination, subordination, or argument relation.\\
 \hline \hline

\multicolumn{2}{|c|}{{\bf Nominal Elements}}\\
\hline
 {\it det} & Determiner.\\
\hline
 {\it case} & Case marker, including adpositions.\\
\hline
 {\it nmod} & Nominal modifier of a noun or a noun phrase.\\
\hline
 {\it nummod} & Numeric modifier.\\
\hline
\end{tabular}
\caption {\label{tab:categories}
Some common UD edge types that are used in this paper, and their definitions.}
\end{table}

\subsection{Major Constructions in UD}\label{sec:ud_cxns}

\paragraph{Auxiliaries and Modals.}
Auxiliary and modal verbs in UD are dependent on the matrix verb. For example, ``can'' in this example is dependent on ``write'':

\enumsentence{
\centering
\raisebox{-.7em}{
\begin{dependency}
 	\begin{deptext}
	 He \& can \& write \& the \& letter \\
 	\end{deptext}
 \deproot[edge unit distance=2.3ex]{3}{root}
 \depedge{3}{2}{aux}
 \depedge{3}{1}{nsubj}
 \depedge{3}{5}{dobj}
 \depedge{5}{4}{det}
 \end{dependency}
}}

\paragraph{Adverbs and Negation.}
Adverbs and negation are treated similarly to auxiliaries and modals, and are also dependents of the matrix predicate.\footnote{See appendix for the transliteration scheme of the Hebrew letters. We adopt the one used in the Hagar corpus.} 

\enumsentence{
\centering
\raisebox{-.7em}{
\begin{dependency}
 	\begin{deptext}
                ha \& sefer \& lo \& niftax \& \textrevglotstop{}a\d{k}\v{s}\={a}yw \\
                the \& book \& not \& open \& now \\
 	\end{deptext}
 \deproot[edge unit distance=2.3ex]{4}{root}
 \depedge{4}{3}{neg}
 \depedge{4}{5}{advmod}
 \depedge{4}{2}{nsubj}
 \depedge{2}{1}{det}
 \end{dependency}
  }
}

\paragraph{Noun Phrases.}
Noun phrases are headed by the lexical head in the case of common NPs, and by the first word in the case of proper nouns. 

\paragraph{Adpositional Phrases.}
Adpositional phrases are represented as dependents of the head noun when found in a noun phrase. When found in a clause, adpositional phrases are represented as dependents of the matrix verb, and are invariably treated as modifiers so as to avoid drawing a hard distinction between core arguments and adjuncts \citep[a difficult distinction to make in practice; see, e.g.,][]{Marc:93}.

\vspace{.5em}
\begin{minipage}{.51\textwidth}
\enumsentence{
\centering
\raisebox{-.7em}{
\hspace{-2mm}
\begin{dependency}
 	\begin{deptext}
	 Press \& the \& button \& on \& the \& chair \\
 	\end{deptext}
 \deproot{1}{root}
 \depedge{1}{3}{dobj}
 \depedge{3}{2}{det}
 \depedge{3}{6}{nmod}
 \depedge{6}{5}{det}
 \depedge{6}{4}{case}
 \end{dependency}
}}
\end{minipage}
\hfill
\begin{minipage}{.45\textwidth}
\enumsentence{
\centering
\raisebox{-.7em}{
\hspace{-2mm}
\begin{dependency}
	\begin{deptext}
	 They \& landed \& on \& this \& spot \\
 	\end{deptext}
 \deproot{2}{root}
 \depedge{2}{1}{nsubj}
 \depedge{2}{5}{nmod}
 \depedge{5}{4}{det}
 \depedge{5}{3}{case}
 \end{dependency}
}}
\end{minipage}

\paragraph{Relative Clauses.}
Relative clauses are internally analyzed just like matrix clauses, where the relative clause's head is considered a dependent of the relativized element. The relative pronoun (where present) is marked with the role of the extracted element. For instance, in the case of object extraction, ``that'' will have a dependency label {\it dobj}:
\enumsentence{
\centering
\raisebox{-.7em}{
\begin{dependency}
 	\begin{deptext}
	 The \& noise \& that \& they \& make \\
 	\end{deptext}
 \deproot{2}{root}
 \depedge{2}{1}{det}
 \depedge{5}{4}{nsubj}
 \depedge{2}{5}{acl:relcl}
 \depedge{5}{3}{dobj}
\end{dependency}
}}

However, where no relative pronoun is present, the extracted slot is underspecified. For instance, ``the noise they make'' and ``the pencil you write with'' are analyzed similarly:

\vspace{.5em}
\hspace{-2em}
\begin{minipage}{.51\textwidth}
\enumsentence{
\centering
\raisebox{-.7em}{
\hspace{-2mm}
\begin{dependency}
 	\begin{deptext}
	 The \& noise \& they \& make \\
 	\end{deptext}
 \deproot[edge unit distance=2.3ex]{2}{root}
 \depedge{2}{1}{det}
 \depedge{2}{4}{acl:relcl}
 \depedge{4}{3}{nsubj}
 \end{dependency}
}}
\end{minipage}
\begin{minipage}{.50\textwidth}
\enumsentence{
\centering
\raisebox{-.7em}{
\hspace{-2mm}
\begin{dependency}
 	\begin{deptext}
		The \& pencil \& you \& write \& with\\
     \end{deptext}
 \deproot[edge unit distance=2.3ex]{2}{root}
 \depedge{2}{1}{det}
 \depedge{4}{3}{nsubj}
 \depedge{2}{4}{acl:relcl}
 \depedge{4}{5}{nmod}
 \end{dependency}
}}
\end{minipage}

\vspace{.1cm}
We therefore introduce two subtypes for the {\it acl:relcl} dependency label: {\it acl:relcl\_subj} and {\it acl:relcl\_obj} for subject and object relative clauses respectively. Where the extracted element is not the subject or the object, we keep the category {\it acl:relcl}, for instance in the case of adjuncts (e.g., ``the pencil you write with'') or extraction from a complement clause (e.g., ``the cat I was taught to like''). The subtyping could be further extended to specify the role of the head noun in those cases, but their frequency in our corpora did not merit further subtyping.

\paragraph{Coordination.}
UD's convention for coordination designates 
the headword of the first conjunct as the head (the other conjuncts are dependent on it with a {\it conj}-labeled edge), while the coordinating conjunctions are dependent 
on the conjunct following them with a {\it cc}-labeled edge.

\enumsentence{
\centering
\raisebox{-.7em}{
\begin{dependency}
 	\begin{deptext}
	 	\textglotstop{}ax\={a}lti \& tap\={u}ax \& we \& \textglotstop{}ag\={a}s \\
                I-ate \& apple \& and \& pear \\
 	\end{deptext}
 \deproot[edge unit distance=2.3ex]{1}{root}
 \depedge{1}{2}{dobj}
 \depedge{2}{4}{conj}
 \depedge{4}{3}{cc}
 \end{dependency}
}}

\paragraph{Open Clausal Complements.}
An open clausal complement of a verb or an adjective (marked as {\it xcomp}) is defined in UD to be a predicative or clausal complement without its own subject.
That is, the subject is inherited from some fixed argument position, often a subject or an object of a higher-level clause. 
Note that raising and control, which differ in the semantic valency of the matrix verb, are not distinguished in the UD parse.

\vspace{1em}
\hspace{-3em}
\begin{minipage}{.4\textwidth}
\enumsentence{
\centering
\raisebox{-.7em}{
\hspace{-5mm}
\begin{tabular}{c}
\begin{dependency}
 	\begin{deptext}
	 I \& want \& to \& sit \\
 	\end{deptext}
 \deproot[edge unit distance=2.3ex]{2}{root}
 \depedge{2}{1}{nsubj}
 \depedge{2}{4}{xcomp}
 \depedge{4}{3}{mark}
 \end{dependency} \\
\textsc{subject control}
\end{tabular}
}}
\end{minipage}
\hspace{-4em}
\begin{minipage}{.4\textwidth}
\enumsentence{
\centering
\raisebox{-.7em}{
\hspace{-5mm}
\begin{tabular}{c}
\begin{dependency}
 	\begin{deptext}
	 I \& want \& him \& to \& sit \\
 	\end{deptext}
 \deproot[edge unit distance=2.3ex]{2}{root}
 \depedge{2}{1}{nsubj}
 \depedge{2}{3}{dobj}
 \depedge[edge unit distance=2ex]{2}{5}{xcomp}
 \depedge{5}{4}{mark}
 \end{dependency} \\
\textsc{raising to object}
\end{tabular}
}}
\end{minipage}
\hspace{-4em}
\begin{minipage}{.4\textwidth}
\enumsentence{
\centering
\raisebox{-.7em}{
\hspace{-5mm}
\begin{tabular}{c}
\begin{dependency}
 	\begin{deptext}
		I \& asked \& him \& to \& sit \\
     \end{deptext}
\deproot[edge unit distance=2.3ex]{2}{root}
 \depedge{2}{1}{nsubj}
 \depedge{2}{3}{dobj}
 \depedge[edge unit distance=2ex]{2}{5}{xcomp}
 \depedge{5}{4}{mark}
 \end{dependency} \\
\textsc{object control}
\end{tabular}
}}
\end{minipage}

\paragraph{Parataxis.}
Where an utterance consists of several clauses or fragments which are not linked through coordination or subordination, but are somewhat loosely related, UD marks the dependency between them as {\it parataxis}. For example:

\enumsentence{
\centering
\raisebox{-.7em}{
\begin{dependency}
 	\begin{deptext}
		Two \& seals \& one \& strong \& man \\
  \end{deptext}
\deproot{2}{root}
 \depedge{2}{1}{nummod}
 \depedge{2}{5}{parataxis}
 \depedge{5}{3}{nummod}
 \depedge{5}{4}{amod}
 \end{dependency}
}}

\paragraph{Ellipsis and Promotion.} 
Where the head word of a phrase is elided, UD's policy is to ``promote'' one of its
children to be the headword. For example, in Example \ref{ex:winston},\footnote{This is an authentic example from the Adam corpus, in which ellipsis is not common.}
the auxiliary ``should''--which would normally serve as a modifier of the matrix clause--instead serves as the head of the adverbial clause. UD does not distinguish between a promoted head and a regular head.

\enumsentence{\label{ex:winston}
\centering
\raisebox{-.7em}{
\begin{dependency}
 	\begin{deptext}[column sep=.7em]
		Winston \& tastes \& good \& like \& a \& cigarette \& should \\
  \end{deptext}
\deproot{2}{root}
 \depedge{2}{1}{nsubj}
 \depedge{2}{3}{xcomp}
 \depedge[edge unit distance=1.7ex]{2}{7}{advcl}
 \depedge[edge unit distance=2ex]{7}{4}{mark}
 \depedge{7}{6}{nsubj}
 \depedge{6}{5}{det}
  \end{dependency}
}}

In order to make this distinction explicit, we subcategorize the dependency label of the promoted word's incoming edge to indicate  that it was promoted to that position (in the above case, {\it advcl:promoted}). We only target VP ellipsis, due to its importance in theories of the syntax-semantics interface, but similar subcategorizations are in principle possible for other elliptical constructions.

\subsection{Constructions Idiosyncratic to CDS}\label{sec:ud_in_cds}

New genres frequently impose new demands on UD annotation guidelines \citep[as can be seen, for example, in the literature on UD for user-generated content;][]{sanguinetti-20}.
We turn to discussing a number of common phenomena from our corpus that are not often found in other UD corpora for English and Hebrew, which mostly target news and web texts. Indeed, there is little UD-annotated data of spoken English (mostly parliamentary proceedings), and none for spoken Hebrew. Our corpora are thus different from most existing corpora in targeting spoken language, and in addressing the specific register of CDS.

\paragraph{Serial Verb Constructions.}
Serial verb constructions (SVCs) are very restricted in English and Hebrew, but are fairly common in CDS.  Examples in Adam only include the verb ``go'' in the first position (e.g.~{\it Go get Hans.}).\footnote{This construction is sometimes referred to as quasi-SVC. See \citep{pullum1990constraints} for discussion.} Examples
in Hagar are semantically similar, but include a somewhat broader class of verbs, such as
``b\={o}\textglotstop{}i tir\textglotstop\={\i}'' (lit.~{\it come see}). 
In the absence of clear UD guidelines as to how to treat this construction, we adopt the UDv2 sub-type for SVCs, {\it compound:svc}, and apply it to this case.
For example:

\enumsentence{
\centering
\raisebox{-.7em}{
\begin{dependency}
 	\begin{deptext}[column sep=2em]
		Go \& get \& Hans \\
  \end{deptext}
 \deproot[edge unit distance=2.3ex]{1}{root}
 \depedge{1}{2}{compound:svc}
 \depedge[edge unit distance=3ex]{1}{3}{dobj}
  \end{dependency}
}}

 \paragraph{Ambiguous Fragments.} Many utterances do not constitute a complete clause, but only parts of it. In some cases, the syntax of such fragments
  may be underspecified. Examples include ``frighten me for" from Adam, where it's unclear what the attachment of ``for'' is, and the following example from Hagar, where the role of ``sgul\={\i}m'' (``purple") is not clear:
  \begin{description}
      \item Sgul\={\i}m  , Hag\={a}ri , an\={\i} lo\textglotstop~ro\textglotstop\={a}
      \item Purple$_{\textit{pl}}$, Hagari, I \hspace{0.3cm} not see$_{\textit{1pl,fem,past}}$
  \end{description}

  In these cases, we instructed annotators to guess to the best of their ability what the sentence might mean and annotate it accordingly.

 \paragraph{In-situ WH-pronouns.} While the grammar of both English and Hebrew requires that 	wh-pronouns ordinarily be fronted in questions, it is quite common to find in Adam cases where 
   the pronouns stay in place. Examples: ``A bird what?'', ``Jiminy Cricket who?'', ``do 	not what?''. The phenomenon occurs in Hagar as well, albeit less commonly.
   We annotate in-situ WH-pronouns the same as we annotate fronted wh-pronouns.

 \paragraph{Word Plays.}
    Some phrases and utterances appear to be playful manipulations of 
    existing words, or belong to some private language
    between the adult and the child. It is not straightforward to 
    determine what the propositional content of such
    cases is, if any. Examples include ``romper bomper stomper boo'' and 
    ``sorbalador'' from Adam and ``balad\={o}n'' and ``bdibiyabi'' 
    from Hagar. Where the invented word is embedded within an 
    otherwise intelligible utterance, annotators are instructed to infer its syntactic 
    category from context. 
    Where the syntax is unclear, we use the residual POS tag X 
    and edge type {\it dep}.
    In such cases, our converter produces no LF for the utterance.

 \paragraph{Non-standard Vocabulary.} Other than word plays,
      examples of non-standard vocabulary include real words or phrases, used
      in a non-standard way.
      For example, ``n\={u}ma 	n\={u}ma'' means ``sleep sleep'' in Hebrew and is part of
      a nursery rhyme. In Hagar, it appears in ``na\textrevglotstop{}a\d{s}\={e} le ha 
      \textglotstop{}efr\={o}ax n\={u}ma n\={u}ma'', which translates to ``we will do to the 
      chick n\={u}ma n\={u}ma'', probably meaning they will put it to bed.
      Other examples may be ungrammatical inflections of real words, e.g.,
      ``play games? boat somes'', where ``boat somes'' probably means ``some boats''.
      We instruct annotators to assign edge labels to words according to their syntactic
      function, rather than according to their standard function in the target language.
      For example, ``n\={u}ma n\={u}ma'' will be considered a direct object in this case,
      despite being a verb morphologically.

 \paragraph{Quotations.} We have observed many examples of utterances including quoted fragments, for
      instance the adult repeating what the child had said, or quoting rhymes, songs, and onomatopoeia.
      Sentences including quotes are not straightforward to analyze syntactically, and even more
      difficult to provide semantic representation for. Examples: ``Adam, can you say sits in the chair
      the boy?'', ``It says gobble gobble'', ``There's a dot that says cross your printing set.'',
      ``Did you say fright or did you say fight?''. We annotate quotations that do not contain a clause
      as direct objects, while quotations that do are annotated as complement clauses.

  \paragraph{Repetitions.} Repetitions of a word or a phrase are common in CDS \citep{Newp:77,hoff1985some}.
      The two major sub-classes are discursive repetitions (``no no don't do that'';
      ``b\=o\textglotstop{}i b\=o\textglotstop{}i'' lit.~``come come'')
      and onomatopoeias (``oink oink''; ``tuk tuk'' which is Hebrew for ``knock knock'').
      Some repetitions elaborate on the first occurrence (``Adam's Adam's what?'';``\d{t}ip\=a, \d{t}ip\=a \v{s}el m\={a}yim'' lit.~``drop, drop of water'') or only partially repeat it (``ma \textrevglotstop{}o\d{s}\={\i}m po ma \textrevglotstop{}o\d{s}\={\i}m'' gloss: ``what do$_{\textit{pl,masc,pres}}$ here what do$_{\textit{pl,masc,pres}}$'', translation: ``what does one do here, what does one do?''). 
     The motivation for some repetitions is obscure, even in
  context (``guess he means ride buggy buggy'').

 We introduce the subtype {\it parataxis:repeat} to indicate repetitions, 
 except in cases where the repetition is
 constructional, as in ``hold your hand way way up'',
 where the repetition is interpreted as an intensifier, and so both ``way'' instances are annotated as {\it advmod}.

\enumsentence{
\centering
\raisebox{-.7em}{
\begin{dependency}
 	\begin{deptext}[column sep=1.6em]
		ha \& sand\={a}l \& qa\d{t}\={a}n \& qa\d{t}\={a}n \& we \& Hag\={a}r \& gdol\={a} \& gdol\={a} \\
        the \& {\it sandal} \& {\it small} \& {\it small} \& {\it and} \& {\it Hagar} \& {\it big} \& {\it big} \\
  \end{deptext}
\deproot{3}{root}
 \depedge{3}{2}{nsubj}
 \depedge[edge unit distance=2ex]{3}{7}{conj}
 \depedge{2}{1}{det}
 \depedge{3}{4}{parataxis:repeat}
 \depedge{7}{5}{cc}
 \depedge{7}{6}{nsubj}
 \depedge{7}{8}{parataxis:repeat}
 \end{dependency} 
}} 

Note that {\it parataxis:repeat} is different than the UD subtype {\it compound:redup}, common in some languages, which denotes the result of the morphosyntactic operation of reduplication.

\section{Converting Dependency Structures to Logical Forms}\label{sec:ud-lf}

\begin{figure*}\centering\small

\begin{tikzpicture}[
blue/.style={rectangle, draw=blue!60, fill=blue!5, very thick, minimum size=7mm, font=\ttfamily},
]
	\node[blue](t) at (2,8) {t/think-01};
	\node[blue](y) at (0,6) {y/you};
	\node[blue](p) at (2,6) {p/possible-01};
	\node[blue](a) at (-2,6) {a/amr-unknown};
	\node[blue](w) at (2,4) {w/want-01};
	\node[blue](w2) at (1,2) {w2/whale};
	\node[blue](m) at (3,2) {m/milk};
	\node[blue](b) at (1,0) {b/baby};
	\node[blue](s) at (3,0) {s/some};
	\draw[->, thick] (t.south) -- (y.north) node[midway, above, sloped] {:ARG0\hspace*{3em}};
	\draw[->, thick] (t.south) -- (p.north) node[midway, above, sloped] {:ARG1};
	\draw[->, thick] (p.south) -- (w.north) node[midway, above, sloped] {:ARG1};
	\draw[->, thick] (w.south) -- (w2.north) node[midway, above, sloped] {:ARG0};
	\draw[->, thick] (w2.south) -- (b.north) node[midway, above, sloped] {:mod};
	\draw[->, thick] (w.south) -- (m.north) node[midway, above, sloped] {:ARG1};
	\draw[->, thick] (m.south) -- (s.north) node[midway, above, sloped] {:mod};
	\draw[->, thick] (t.south) -- (a.north) node[midway, above, sloped] {:polarity};
\end{tikzpicture}

\vspace{.5cm}
\begin{tikzpicture}[
blue/.style={circle, draw=blue!60, fill=blue!60, very thick, font=\ttfamily},
]

	\node[blue](root) at (4,6) {};
	
	\node[blue](do) at (0,4) {Do};
	\node[blue](you) at (1.5,4) {you};
    \node[blue](think) at (3,4) {think};
	\node[blue](a1) at (4.5,4) {};
	\node[blue](punct1) at (8,4) {?};
	
	\node[blue](babyWhale) at (2,2) {}; 
	\node[blue](might) at (3.5,2) {might};
	\node[blue](want) at (5,2) {want};
    \node[blue](imp) at (6.5,2) {IMPLICIT}; 
    \node[blue](someMilk) at (9,2) {}; 
	
	\node[blue](the) at (-1.5,0) {the};
	\node[blue](baby) at (0,0) {baby};
	\node[blue](whale) at (1.5,0) {whale};
    
    \node[blue](some) at (8,0) {some};
	\node[blue](milk) at (10,0) {milk};	
    
 	\draw[->, thick] (root) -- (do) node[midway, above, sloped] {Function};
 	\draw[->, thick] (root) -- (you) node[midway, above, sloped] {Part.};
 	\draw[->, thick] (root) -- (think) node[midway, above, sloped] {Process};
 	\draw[->, thick] (root) -- (a1) node[midway, above, sloped] {Part.};
 	\draw[->, thick] (root) -- (punct1) node[midway, above, sloped] {Punctuation};
 	 \draw[->, thick] (babyWhale) -- (the) node[midway, above, sloped] {Function};
   \draw[->, thick] (babyWhale) -- (baby) node[midway, above, sloped] {Elab.};
   \draw[->, thick] (babyWhale) -- (whale) node[midway, above, sloped] {Center};
 	 \draw[->, thick] (a1) -- (babyWhale) node[midway, above, sloped] {Part.};
   	 \draw[->, thick] (a1) -- (might) 
     node[midway, above, sloped] {Adv.};
     	 \draw[->, thick] (a1) -- (want) node[midway, above, sloped] {Adv.};
    \draw[->, thick] (a1) -- (imp) node[midway, above, sloped] {Process};
    \draw[->, thick] (a1) -- (someMilk)     
         node[midway, above, sloped] {Part.};
    \draw[->, thick] (someMilk) -- (some)     
         node[midway, above, sloped] {Quant.};
    \draw[->, thick] (someMilk) -- (milk)     
         node[midway, above, sloped] {Center};
\end{tikzpicture}
    \caption{Example AMR (top) and UCCA (bottom) graphs for the sentence ``Do you think the baby whale might want some milk?'' Abbreviations: Part. (Participant), Elab. (Elaborator), Quant. (Quantity) and Adv. (Adverbial).}
    \label{ex:amr_ucca}
\end{figure*}

The purpose of the system presented in this section is to generate semantic representations on the basis of syntactic ones in a way that is automatic and cross-linguistically applicable. The syntactic representation assumed as the input is in the form of UD, complete with Universal POS tags for each word.

The logical forms we use focus on \nssmod{\emph{compositional sentential semantics} -- in particular,} argument structure phenomena. An example for the sentence ``Do you think the baby whale might want some milk ?'' is as follows:
\begin{center}
\lf{ \lamnest{1}{Q(do}{think}{you, \lamnest{2}{might}{want}{\\
\quant{the}{x}{and\_comp(baby(x), whale(x))}, \quant{some}{y}{milk(y)}}}}
\end{center}

\nssmod{Much of this notation should be familiar as a standard Neo-Davidsonian approach to logical semantics, expressed by lambda forms. Briefly, this LF uses two event variables $e_1$ and $e_2$, one for {\it think} and one for {\it might want}.  These are introduced by $\lambda$ terms and notated as subscripts on predicates associated with the event. The utterance is a polar question, denoted by $Q$. Two entity variables, $x$ and $y$, are respectively introduced by the generalized quantifiers \textsc{the} and \textsc{some}. Most content concepts are represented as semantic predicates with names derived from words in the sentence.}

\oamod{\nssmod{With a focus on predicate-argument structure, the LFs} are similar in their core semantic content to other broad-coverage semantic schemes, such as AMR 
\citep{amr} and UCCA \citep{abend2013universal}.
For comparison, Figure~\ref{ex:amr_ucca} presents the above sentence represented as a UCCA graph and an AMR graph. 
All three representations capture the argument structure of the sentence, (semantic) head-dependent relations, and semantic types of the various constants and variables.
AMR and UCCA}\nssmod{ go beyond the LFs in capturing elements of \emph{lexical semantics} (e.g., word senses, semantic roles), as well as \emph{discourse meaning} (e.g., coreference). 
Though it could be valuable to build upon our LFs to incorporate these other aspects of meaning, they are not a part of our investigation here.\footnote{\label{fn:lexical-semantics}\nssmod{That LFs are focused on compositional aspects of meaning is what allows us to induce them from UD trees without relying on additional resources such as lexicons. The LFs shallowly equate content words with concepts; they do not incorporate word sense disambiguation or semantic roles, but do use \oamod{POS tags, morphological inflection features and} multiword expressions to the extent encoded in UD (\S\ref{sec:mwes}) for disambiguation purposes.
}}}

The LFs do offer some advantages over some of the aforementioned alternatives. First, they offer a straightforward decomposition into sub-parts that align with individual words. This is in contrast to schemes, like AMR, that do not offer such a decomposition \citep{szubert-etal-2018-structured}.
This property of the LFs is useful for modeling or evaluating compositionality in the context of child language learning (see \S\ref{sec:simulations}).
Second, the LFs can be transduced using a flexible framework (detailed below), that can easily incorporate further (or fewer) distinctions, if provided with the relevant features in the input. 

These considerations motivate our choosing LFs over other semantic schemes of semantic representation.
\oamod{We further note that the LFs reflect an underspecified approach to representation that is in line with (i.e., does not make any modeling decisions that contradict)  more elaborate semantics that can be applied to these sentences, such as lexical semantics or quantifier semantics. 
However, we do not see the LFs as superior to other alternatives, and note that a similar resource and analysis could have been produced with other schemes as well.}

The output Logical Forms (LF) are typed lambda calculus expressions, and the theoretical approach to semantic representation broadly follows the event semantics of  \citet{Davi:67}.
Our system is based on UDepLambda of \citep{reddy2016transforming}, which we modified to accommodate a different target LF. We stress that the LFs do not contain lexical semantic information about the words involved, and the transcribed words themselves are generally used as their logical constants (e.g., ``pencil'' and ``blue'' are used in Figure~\ref{fig:tt_lfa} to refer to the concept of a pencil and the color blue).

UDepLambda is a conversion system based on the assumption that Universal Dependencies can serve as a scaffolding for a compositional semantic structure -- individual words and dependency relations are assigned their semantic representations, and those are then iteratively combined to yield the 
representation of the whole sentence. 
Our modification to UDepLambda consists of providing a new set of rules, which defines a semantics different from the default one used by UDepLambda.

In what follows we present the UD-to-LF conversion process and discuss our choice of LF.

\subsection{Conversion process}\label{sec:conversion}

Converting a UD parse to an LF is a three-stage process:

\begin{figure}
	\centering
    \includegraphics[width=\textwidth]{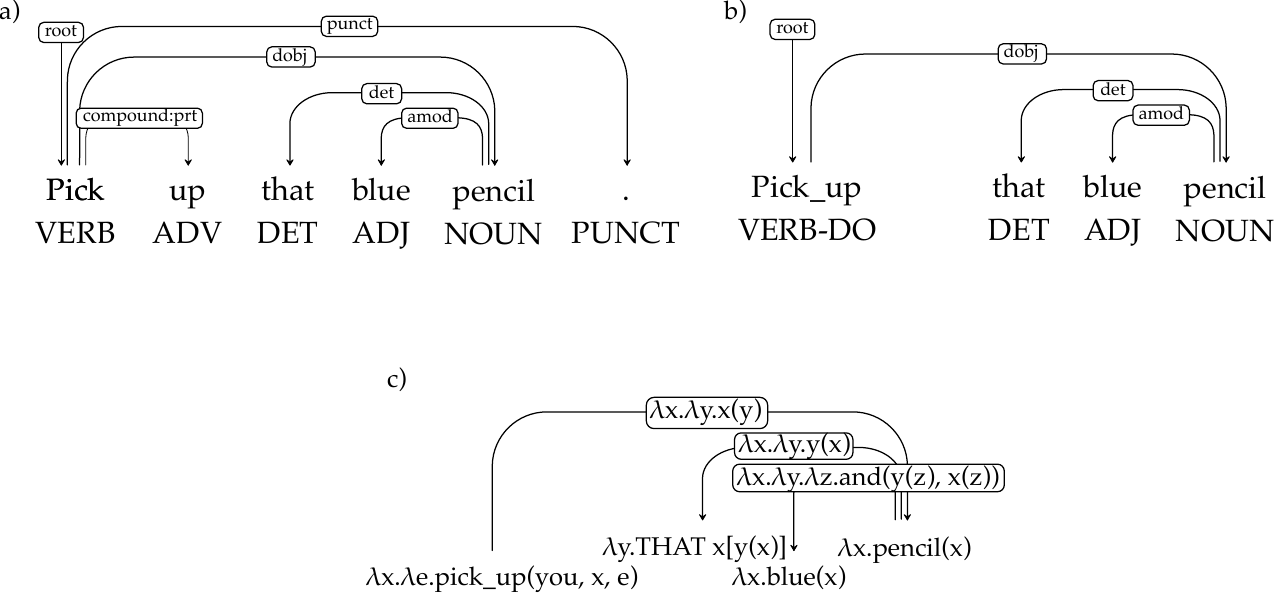}
    \caption{a: UD parse, b: tree transformation to subcategorize verb POS, remove punctuation, and combine verb with its particle, c: LF assignment to nodes and edges.}
    \label{fig:tt_lfa}
\end{figure}

\begin{figure}
\centering
\includegraphics[height=.9\textheight]{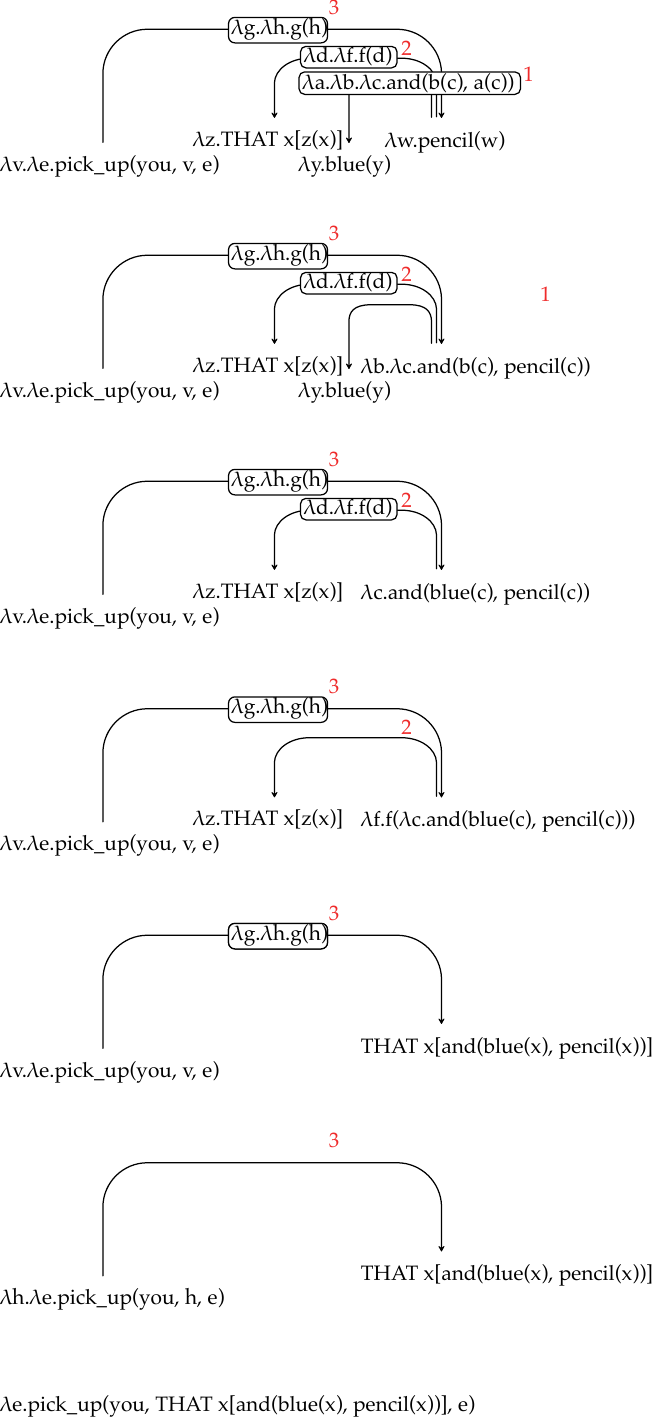}
\caption{Derivation of the LF for the sentence \textit{Pick up that blue pencil}, starting after $\alpha$-conversion of the LF expressions. Reduction proceeds by applying the LF of the dependency relation to the LF of the head, and applying the resulting LF to the LF of the dependent. The red numbers mark the order of composition determined in the tree binarization step.}
\label{fig:deriv}
\end{figure}

\begin{itemize}
\item Tree transformation: as an initial step of conversion we modify the parse trees in order to facilitate the subsequent process of LF assignment. The transformations primarily include subcategorizing POS and dependency labels and removing semantically vacuous items. The rules used in this process (as well as LF assignment rules) consist of a tree regular expression \citep[Tregex;][]{levy-06} and an action to be taken when the pattern is matched. The example in Figure~\ref{fig:tt_lfa}(b) illustrates subcategorization of the POS tag of a verb whose only core argument is a direct object. A tregex is used which matches a verb with an outgoing {\it dobj, ccomp} or {\it xcomp} dependency but without a {\it nsubj} or {\it iobj}, and not in a subject control context (i.e.~with an incoming {\it xcomp} edge); when a node is a match, we change its POS label to VERB-DO.
Most transformation rules depend only on the syntactic context (POS tags and dependency labels), with the only exception being the lexicalized rules for recognizing question words. There are 120 rules in total.

\item LF assignment: Each dependency and each lexical item in the sentence are assigned a logical form, based on their POS tag / edge label and their syntactic context, as in Figure~\ref{fig:tt_lfa}(c). The LF assignment rules are not lexicalized. There are 230 assignment rules. \aumod{For simplicity of presentation here, we write the logical constants in the LFs in the same way as their corresponding words. However, in the corpus the logical constants indicate the POS, lemma and inflection given in the CHILDES annotations. For example, the constant corresponding to the base form of the verb {\it think} would be {\it verb$|$think}, while {\it thinking} would be {\it participle$|$think-presentProgressive}. These symbols are treated atomically by the converter, so they serve as a way to minimally disambiguate different inflections with the same surface form, but otherwise the POS and morphology are not used by the converter.}

\item Tree binarization and LF reduction: The parse tree is binarized to fix the order of composition of word- and dependency-level LFs. Binarization follows a manually created list of dependency priorities. With the order fixed, the sentence-level LF is obtained through beta-reduction, as shown in Figure~\ref{fig:deriv}.
\end{itemize}

All rules used in the conversion process are manually created and assigned priorities. UD trees are processed top-to-bottom and the first transformation and LF assignment rule which matches a given node or edge is applied.

Introducing subcategorizations at the tree transformation step is largely a matter of convenience. The same distinctions could in principle be encoded in LF assignment rules. However, introducing more fine-grained labels makes LF assignment rules easier to write and maintain.

\subsection{Target Logical Forms}\label{sec:lfs}

Our target is a Davidsonian-style event semantics, encoded in a typed lambda calculus.\footnote{This paper does not focus on the type system, but roughly speaking, it works as follows. There are three base types in our calculus: \textit{t} for truth type, \textit{v} for variable (individual) type, and \textit{r} for event type. Predicates with all argument slots saturated are functions from events to truth value, hence type \textit{$<$r,t$>$}. In the target LFs only variables and constants are typed, but during derivation all expressions are typed.} In this section we describe the output we designed for the converter without claiming it to be ``target semantics'' understood as an ideal meaning representation.

An utterance is assumed to describe an event, and the LFs typically contain an event variable with scope over the whole expression. 
For example, the LF for the sentence \textit{You found it} is  

\begin{center}
	\lf{\lam{1}{found}{you, it}}
\end{center}

In the interest of legibility we show the event variable as a subscript of all predicates it has scope over instead of showing it as their argument.
\aumod{In the corpus all variables are typed, the event variable is always the last argument of the predicate.}

We turn to discussing the resulting representations for a number of common phrase types and constructions. 

\subsubsection*{Nominals}
This category includes common and proper nouns, as well as pronouns.

\noindent Pronouns and proper nouns are treated as referring expressions and are represented as atomic terms. Common nouns are treated as non-referring and represented as functions of arity 1, requiring an argument to become referential. When determiners and quantifiers combine with common nouns, they provide such an argument by introducing a variable which they bind.
Mass nouns and plural nouns, despite not requiring a determiner, are represented the same way as full determiner phrases, with a placeholder \textit{BARE} determiner.\footnote{The actual format in the corpus is of the form \lf{your(x, toy(x))}, \lf{BARE(x, toy(x))}. In the paper, we adopt a format more familiar from literature on quantifier phrase semantics for the sake of readability.}

\begin{description}
	\item it: \lf{it}
	\item Adam: \lf{Adam}
	\item toy: \lf{ \lamn{x}{toy(x)} }
	\item a toy: \lf{ \quant{a}{x}{toy(x)} }
	\item toys: \lf{ \quant{bare}{x}{toy(x)} }
\end{description}

\oamod{\noindent Where proper nouns appear with a determiner, they are treated similarly to common nouns:

\begin{description}
	\item the Daddy: \lf{ \quant{the}{x}{Daddy(x)} }
\end{description}
}

\oamod{\noindent Possessives are treated in the same way as determiners:

\begin{description}
	\item your toy: \lf{ \quant{your}{x}{toy(x)} }
        \item Diandro's bottle: \lf{ \quant{Diandro's}{x}{bottle(x)} }
\end{description}
}

\noindent When used predicatively (notably, in copular constructions), 
nominals are treated as predicates with an arity of 2, taking as arguments a subject and an event variable. All nominals therefore have two possible types of LFs, non-predicative and predicative.

\oamod{Where nominals are used predicatively, we do not interpret their determiner as having a determiner semantics in the LF, and instead simply interpret it as an application of the predicate defined by the nominal to the subject:

\begin{description}
	\item It is a raccoon: \lf{  \lam{1}{raccoon}{it}}
	\item My pet is a raccoon: 
 \lf{  \lam{1}{raccoon}{\quant{my}{x}{pet(x)}}}
    \item This is the car: 
    \lf{  \lam{1}{the\_car}{this}}
\end{description} 

}

\noindent Noun-noun compounds are represented by treating both nouns as arguments to a special \textit{and\_comp} predicate.

\begin{description}
	\item Show me a space boat:\lf{ \lam{1}{show}{you, \quant{a}{x}{and\_comp(space(x), boat(x))}, me} }
\end{description}

\subsubsection*{Adjectives}
Like common nouns, adjectives are represented as arity 1 predicates. We assume intersective semantics for adjectives, 
i.e., a nice carpenter is a thing which is nice and which is a carpenter. 

\begin{description}
	\item nice: \lf{ \lamn{x}{nice(x)} }
	\item nice carpenter: \lf{ \lamn{x}{and(nice(x), carpenter(x))} }
\end{description}
This is decidedly a simplification of the actual nuanced adjective semantics, e.g., a fake bear is not really a bear, and a good liar is not a person who is good and who is a liar.

Adjectives can head copular constructions, in which case they behave like arity 2 predicates, analogously to nominals in the same situation. 
It is possible for an adjective in those constructions to also have a clausal object, increasing arity to 3.

\begin{description}
	\item This carpenter was nice: \lf{ \lam{1}{nice}{\quant{this}{x}{carpenter(x)}} }
	\item I am sorry to go: \lf{ \lam {1} {sorry} {I, \lam{2}{go}{I}} }
\end{description}

\subsubsection*{Verbs}
Verbs are represented by predicates whose arity varies from 1 to 4, with possible arguments being subject, direct object, indirect object, \oamod{clausal arguments (see below)} and the Davidsonian event variable (represented in that order in the LF). The argument type is defined by its syntactic relation to the verb in the unmarked form. If a verb takes less than 4 arguments, we leave the other positions unfilled.

\begin{description}
	\item You gave Ursula the box:  \lf{\lam{1}{gave}{you, \quant{the}{x}{box(x)}, Ursula}}
	\item Mommy heard it: \lf{\lam{1}{heard}{Mommy, it}}
\end{description}

\noindent When a verb lacks an argument whose position precedes the positions of present arguments (with the exception of the event argument), we fill the slot of the missing argument with a ``blank'' symbol (\_). Constructions necessitating this solution include passive voice\footnote{As argument roles are defined semantically, in a passive clause the syntactic subject becomes one of the objects in the LF, and the subject spot is left empty.} and some infinitival clausal arguments.

\begin{description}
	\item Daddy said to return the pen: \lf { \lam{1} {said} {Daddy, \lam{2} {return} {\_, \quant{the}{x}{pen(x)}}} }
	\item The tree is shaped (like that): \lf{ \lam{1} {shaped} {\_, \quant{the}{x}{tree(x)}} }
\end{description}

\textbf{Subject-less clauses.} For every verb without a subject\footnote{This does not include verbs which have a subject that is not directly connected to the verb in the UD parse -- verbs controlled by a higher verb, verbs in infinitival complement clauses, verbs in relative clauses, verbs sharing a subject with a conjoined verb.} we assume the clause is in imperative mood and supply \textit{you} in the subject position in the LF.

 \begin{description}
 	\item Drink the juice:	\lf{ \lam{1}{drink}{you, \quant{the}{x}{juice(x)}} }
 \end{description}

\textbf{Auxiliary and modal verbs} are predicates with an arity of 1, taking as their argument a proposition.

\begin{description}
	\item He can write: \lf{ \lamnest {1} {can} {write} {he} } 
    \item He could be writing: \lf{ \lamnest{1}{could}{be}{writing$_{e_1}$(he)}}
\end{description}

\textbf{Particle verbs}, including phrasal verbs, are merged into one lexical item of the form \textit{verb\_particle} whenever there are no other words intervening between the verb and its particle. Otherwise the particle is treated as a sentential modifier. The difference in treatment is motivated purely by the technical limitations of the converter, not theoretical considerations.

\begin{description}
	\item The paint came off: \lf{ \lam{1} {came\_off} {\quant{the}{x}{paint(x)}} }
	\item It picks the dirt up: \lf{ \lamand{1} {picks} {it, \quant{the}{x}{dirt(x)}} {up} {} }
\end{description}

\textbf{Serial verb constructions} of the form 
 ``come get'' or ``go ask'' and their Hebrew counterparts (e.g., 
 ``b\={o}\textglotstop{}i te\v{s}v\={\i}'', lit.~``come sit'') are treated in a special way, because semantically the first verb carries little propositional meaning and is purely discoursive in nature.
    Our converter reduces these expressions to the second verb only.
    \begin{description}
    	\item Go get two pennies: \lf{ \lam{1} {get} {you, \quant{two}{x}{pennies(x)}} }
    \end{description}

\subsubsection*{Adverbs}
Verb-modifying adverbs are represented as predicates which take the event variable as their argument, and are conjoined with the matrix predicate using a general purpose \textit{and}. We do not distinguish between VP-scoped and sentential adverbs because this distinction is not supported by the UD annotation.

\begin{description}
	\item She tried again: \lf{ \lamand{1}{tried}{she}{again}{} }
	\item She certainly tried: \lf{ \lamand{1}{tried}{she}{certainly}{} }
\end{description}

\noindent Adjuncts (which are annotated as adverbs) modifying adjectives are arity 1 predicates whose argument is the LF representation of the modified adjective phrase.

\begin{description}
	\item a very kind boy: \lf{ \quant{a}{x}{and(very(kind(x)), boy(x))} }
	\item You are a very kind boy: \lf{ \npred{1} {a} {you} {and(very(kind\textsubscript{\var{1}}(you)), boy} {you)} }
\end{description}

\subsubsection*{Prepositional phrases}
Due to the difficulty in making the complement-adjunct distinction in UD, prepositional phrases (PP) within clauses are invariably
considered as sentential modifiers (rather than arguments).
A preposition is an arity 2 predicate, whose first argument is the prepositional object, and the second
is the event variable, and the LF of the prepositional phrase is conjoined with the LF of the matrix predicate.

\begin{description}
	\item He played with Paul: \lf{ \lamand{1}{played}{he}{with}{(Paul)} }
\end{description}

\noindent A PP modification of a nominal is represented using the \textit{att} relation, expressing the fact that the PP is in some sense an attribute of the nominal.

\begin{description}
	\item the juice on your shirt: \lf{ \quant{the}{x}{att(juice(x), on(\quant{your}{y}{shirt(y)}))} }
\end{description}

\noindent When a PP is used in a copular construction, the preposition is represented by an arity 3 predicate, taking as arguments the nominal inside the PP, the subject, and the event variable.

\begin{description}
	\item That is from Pinocchio: \lf{ \lam{1}{from}{Pinocchio, that} }
\end{description}

\subsubsection*{Relative clauses}
Relative clauses provide additional information about a nominal which they modify. We represent the relation between a nominal and a relative clause as conjunction. There is no difference between LFs of normal and reduced relative clauses, or between restrictive and non-restrictive ones.

\begin{description}
	\item We saw those mirrors that you liked:
	
	\lf{ \lam{1}{saw}{we, \quant{those}{x}{and(mirrors(x), \lam{2}{liked}{you, x})}} }
    
    \item The drum you were playing:
    
    \lf{ \quant{the}{x}{and(drum(x), \lamnest{1}{were}{playing}{you, x})} }
    
\end{description}

Free relative clauses, as in the example below, pose problems to the UD scheme. In absence of clear annotation guidelines, we decided to attach the relative clause to the matrix clause with the \textit{ccomp} or \textit{csubj} relation and annotate the wh-word in a way that reflects its role within the relative clause.
We use whatever relation is appropriate, and subcategorize it with a complementizer subtag, \textit{:comp}.

\begin{figure}[h]
	\centering
	\begin{dependency}
		\begin{deptext}[column sep=1.5em]
			You \& heard \& what \& I \& said \& . \\
		\end{deptext}
		\deproot{2}{root}
		\depedge{2}{1}{nsubj}
		\depedge{5}{3}{dobj:comp}
		\depedge{5}{4}{nsubj}
		\depedge{2}{5}{ccomp}
	\end{dependency}
\end{figure}

Using this annotation convention we can produce correct LFs for fused relative clauses:
 \begin{description}
 	\item You heard what I said: 
 	\lf{ \lam{1}{heard}{you, \quant{what}{x}{\lam{2}{said}{I, x}}} }
 \end{description}

\subsubsection*{Clausal arguments and modifiers}
Clauses can function as arguments of verbs and, less often, other predicates. In LF clausal arguments are treated no different from nominal ones. 

\begin{description}
	\item I think that he can talk: \lf{ \lam{1}{think}{I, \lamnest{2}{can}{talk}{he}} }
	\item He wants you to take a nap: \lf{ \lam{1}{wants}{he, \lam{2}{take}{you, \quant{a}{x}{nap(x)}}} }
\end{description}

Generating LFs for clausal complements is complicated by the ambiguity of the UD scheme which does not distinguish between raising to object and object control constructions. The actual semantics differs, but our converter heuristically treats all open clausal complements as if they were cases of object control and produces the LF accordingly. See discussion in \S\ref{sec:limitations}.

Clausal modification of verbs is represented by treating the matrix clause and the subordinate clause as two arguments of the subordinating conjunction predicate. The predicates representing both clauses share the event variable.\footnote{It may be argued that in some cases, the subordinate clause does not in fact correspond to the same event as the main clause. This is a fine distinction not made by UD, and consequently not made in the produced LFs either.}

\begin{description}
	\item She sings when she is happy: \lf{ \lamconj{when} {1} {happy} {she} {sings} {(she)} }
\end{description}

\noindent Clausal modifiers of nominals (other than relative clauses) come in two types. The first have relative clause semantics:
 \begin{description}
 	\item You saw a tree dancing: \lf{ \lam{1}{saw}{you, \quant{a}{x}{and(tree(x), \lam{2}{dancing}{x})}} }
 \end{description}
The second type of modification is more difficult to encode, as the specific semantic relation between the noun and the modifier is largely implicit. Examples include noun phrases such as \textit{a battle to keep him out}, \textit{one place to put things}, or \textit{the way to play}. We resort to representing the relation with a generic \textit{rel} predicate.
 \begin{description}
 	\item You showed me the way to play the game:\\
 	\lf{ \lam{1}{showed}{you, me, \quant{the}{x}{rel(way(x), \lam{2}{play}{\_, \quant{the}{y}{game(y)}})}}}
 \end{description}

\subsubsection*{Negation}
Negation of the main predicate of a clause is represented by arity 2 \textit{not} predicate, taking as arguments the negated predicate applied to its arguments and the event variable. The LF of negated nominals follows UD in treating the negation as a determiner.
We note that the auxiliary verb ``do'' is introduced into the logical form whenever it appears as a word in the sentence. Its role in the LF is to serve as a placeholder for tense.

 \begin{description}
 	\item I don't have any sugar: \lf{ \lamnest {1} {not\textsubscript{\var{1}}(do} {have} {I, \quant{any}{x}{sugar(x)}})}
 	\item I'm no clown: \lf{ \npred{1} {no} {I} {clown} {I} }
 \end{description}

\subsubsection*{Questions}
Polar questions are represented by wrapping the LF of the corresponding indicative sentences in a \textit{Q} predicate of arity 1.

 \begin{description}
 	\item Do you have a doll?: \lf{ \lamnest{1}{Q(do}{have}{you, \quant{a}{x}{doll(x)})} } 
 \end{description}

\noindent Wh-questions are represented by binding in the outer scope the variable which stands for the thing being asked about. This variable is used in the LF in place of the wh-word.

\begin{description}
 	\item 
  What did you take?: \lf{ \lamwh{x}{\lamnest{1}{did}{take}{you, x}} }
\end{description}

\oamod{Since possessive modifiers are treated the same as quantifiers, we interpret \textit{whose} questions as abstracting over a generalized quantifier, as in the following example (where we have replaced the variable $x$ with {\sc whose}, in the interest of clarity):
}

\begin{description}
 	\item 
  Whose name are you writing?: \lf{ \lamwh{\textsc{whose}}{\lamnest{1}{are}{writing}{you, \quant{whose}{y}{name(y)}}} } 
\end{description}

\subsubsection*{Conjunctions}
Conjunctions are represented by treating the conjuncts as arguments of the conjunction predicate.\footnote{Two expressions can be conjoined only if they are of the same semantic type.} In cases of clause conjunction there is only one event variable with scope over both clauses.

 \begin{description}
 	\item He had a fever or a cold:\\
  \lf{ \lam{1}{had}{he, or(\quant{a}{x}{fever(x)}, \quant{a}{y}{cold(y)})} }
 	
 	\item 
  \oamod{\textglotstop{}ax\={a}lti tap\={u}ax(x) we \textglotstop{}ag\={a}s (lit. I-ate apple and pear):}\\
  \lf{ \lam{1}{\textglotstop{}ax \={a}lti}{1sg, and(\lamn{x}{tap\={u}ax(x)}, \lamn{y}{\textglotstop{}ag\={a}s(y)})} }
 	\item Get a kleenex and wipe your mouth:\\
 	\lf{ \lamand{1}{get}{you, \quant{a}{x}{kleenex(x)}}{wipe}{(you, \quant{your}{y}{mouth(y)}}) }
 \end{description}

\noindent Shared arguments of conjoined verbs\footnote{There are only a few examples of conjoined predicates with shared arguments in the Adam corpus, but the converter can deal with cases more complicated than the one shown, such as conjoined verbs in relative clauses with the head noun being an argument of both (e.g.~\textit{I'll show you the book I wrote and he edited}).} are explicitly repeated in the LF, as if they were overtly repeated in the sentence. We use a heuristic rule to decide whether an argument is shared or not, which we further discuss in \S\ref{sec:limitations}.

 \begin{description}
 	\item You find and bring it: \lf{ \lamand{1}{find}{you, it}{bring}{(you, it)} }
 \end{description}

\subsubsection*{Names and multiword expressions}\label{sec:mwes}
We combine words annotated with the \textit{mwe}, \textit{name} and \textit{goeswith} relations into a single lexical item and treat them as such in the LF. These three categories are used in UD to classify \nssmod{a restricted subset of} multiword expressions: \textit{name} connects the words of \nssmod{headless names}; \textit{mwe} connects fixed grammaticized expressions; \textit{goeswith} connects two parts of the same word \nssmod{that are incorrectly rendered as separate tokens.\footnote{\url{https://universaldependencies.org/docsv1/u/dep/index.html}}
Many other multiword expressions are syntactically (semi-)regular but semantically idiomatic; our LFs do not capture these as single concepts, since doing so would, at present, require additional layers of annotation (or language-specific lexical resources and disambiguation).
However, other efforts are underway to accommodate a broader range of multiword expressions within the UD framework \citep{parseme-ud-23}, which will in turn enable their treatment by the converter.}

\subsubsection*{Parataxis and discourse markers}
The loose semantic connection associated with the UD relations of {\it discourse} and {\it parataxis} is represented by conjoining the LFs of both parts with a general purpose \textit{and} predicate.

 \begin{description}
 	\item Wait, we forgot your snack: \lf{ \lamand{1}{wait}{you}{forgot}{(we, \quant{your}{x}{snack(x)})} }
    \item I like it, thank you: \lf{ \lamand{1}{like}{I, it}{thank\_you}{} }
 \end{description}

\subsubsection*{Repetitions}
 When repetition annotated with {\it parataxis:repeat} occurs, the LF ignores the repeated element and represents it only once.

    \begin{description}
    	\item \d{t}ip\=a, \d{t}ip\=a \v{s}el m\={a}yim (lit.~drop, drop of water):  \lf{ \quant{bare}{x}{att(\d{t}ip\=a(x), \v{s}el(\quant{bare}{y}{m\={a}yim(y)}))} }
    \end{description}

\subsection{Limitations}\label{sec:limitations}

\nssmod{As observed throughout this section, our LFs encode compositional sentential semantics. The representation does not aim to capture aspects of meaning in the realm of lexical semantics or discourse.

Even in the realm of compositional semantics,} there are cases when the information available from the parse tree and POS tags is not sufficient to recover the correct LF. Most of the limitations of the converter have to do with shortcomings of UD as a syntactic annotation schema, discussed in \S\ref{sec:UD}. In cases of unresolvable structural ambiguity we generally choose to use the LF that represents the meaning which is more common in our English corpus. \oamod{In a number of cases (listed below), where there is no such obvious more common option, we design the converter to simply fail.}

Universal Dependencies builds its structures directly on the words of the sentences, and 
generally does not encode implicit elements or long-range dependencies. 
This often entails difficulties for our conversion method, in cases such as imperatives (e.g., ``come over!''),
where the person addressed is not explicit.
Enhanced Universal Dependencies \citep{Schuster2016enhanced} extend UD and construct
graphs over the input tokens (rather than trees), that cover phenomena such as predicate ellipsis (e.g., gapping),
and shared arguments due to coordination, control, raising and relative clauses.
However, they do not address a variety of implicit arguments, even constructional ones, such as the person
addressed in imperative forms.
We do not use Enhanced UD in this work due to its language-specificity and use of non-tree structures, which considerably complicate
the conversion method.

\subsubsection*{Scope ambiguity}
The major source of ambiguity when deriving an LF from a UD parse is UD's inability to represent scope phenomena. UD trees are not binary and contain no indication about order of composition of the children with the parent, which gives rise to various cases of unclear scope. This is an inevitable consequence of using dependency grammar as annotation rather than (for example) CCG \citep{ccg}.

\begin{itemize}
	\item \textbf{Argument sharing and modifier scope in verb coordination.}
    Coordination structures are inherently ambiguous in UD, as the headword of the first conjunct serves also as the head of the entire coordination structure \citep[for attempts to enhance UD with more informative annotation of coordination structures, see, e.g., ][]{przepiorkowski-patejuk-2019-nested,grunewald-etal-2021-coordinate}. Arguments of the first conjunct and of the whole coordination structure are rendered indistinguishable. The same holds for modifiers of the first conjunct.
For example, in

\begin{figure}[h]
\centering
\begin{dependency}
 	\begin{deptext}[column sep=1em]
	 	Yesterday \& you \& saw \& a \& clown \& and \& ran \\
 	\end{deptext}
 \deproot{3}{root}
 \depedge[edge unit distance=2.7ex]{3}{1}{advmod}
 \depedge{3}{2}{nsubj}
 \depedge{3}{5}{dobj}
 \depedge{5}{4}{det}
 \depedge{7}{6}{cc}
 \depedge[edge unit distance=2ex]{3}{7}{conj}
 \end{dependency}
\end{figure}

it is unclear whether ``clown'' is an object of ``saw`` and ``ran``, or just of ``saw'', and whether both actions or just one happened yesterday.

    The heuristic we select is: if the head verb has an argument which the other verb lacks, assume that the argument is shared. This leads us to correctly represent sentences such as:

 \begin{description}
 	\item You find and bring it: \lf{ \lamand{1}{find}{you, it}{bring}{(you, it)} }
 \end{description}
 
 but also to produce some erroneous LFs:
  \begin{description}
  	\item You saw a clown and ran:
  	
  	\textbf{is} \lf{ \lamand{1}{saw}{you, \quant{a}{x}{clown(x)}}{ran}{(you, \quant{a}{x}{clown(x)})} }
  	
  	\textbf{should be} \lf{ \lamand{1}{saw}{you, \quant{a}{x}{clown(x)}}{ran}{(you)} }
  	
  \end{description}
   
  With respect to modification, we assume that all modifiers attached to the first verb modify the whole conjunction:

  \begin{description}
 	\item She ate and drank again: \lf{ \lamand{1}{and(ate\textsubscript{\var{1}}(she), drank}{she)}{again}{} }
 \end{description}

  In principle the order of words in the sentence could be used for disambiguation: in English shared objects would occur after the second conjuncts, while objects belonging only to the first conjunct would follow it directly. This, however, would require us to provide the converter with linear order information in addition to the UD parse, and make the converter language-specific.

\item \textbf{Modal verb scope.}
  UD treats auxiliaries and modals as modifiers of the matrix verb, giving rise to ambiguity in coordinate structures\footnote{For example, the modal verb applies to the first conjunct in \textit{I will eat the banana but I prefer apples}, but to both conjuncts in \textit{You will sing and play}. Nevertheless, UD attaches the modal verb in both cases to the first conjunct.} and ambiguity over the order of combination of modals and adverbs. The source of the difficulty is the lack of distinction between sentence and VP adverbials in UD. We heuristically decide to represent modals as always outscoping adverbs, correctly representing examples such as this:
  \begin{description}
 	\item Somebody will stop suddenly: 
 	
 	\lf{ \lamnest{1}{will}{and(stop}{somebody), suddenly\textsubscript{\var{1}}} }
 \end{description}
 
 but not others:
 \begin{description}
 	\item Maybe somebody will stop:
 	
 	\textbf{is}	\lf{ \lamnest{1}{will}{and(stop}{somebody), maybe\textsubscript{\var{1}}} }
 	
 	\textbf{should be}	\lf{ \lamand{1}{will\textsubscript{\var{1}}(stop}{somebody)}{maybe}{} }
 	
    Similarly as discussed above, a solution could be proposed which would rely on distinguishing between sentential and VP adverbials on the basis of word order, but this information is not available to the converter.
 	
 \end{description}

\item \textbf{Modifier scope in NP coordination.}
Analogously to the ambiguity arising in verb coordination structures, any modifiers attached to the head noun of a noun coordination structure cause ambiguity. We choose to treat all modifiers as applying to the head of the conjunction only. This results in correct LFs for sentences such as:

 \begin{description}
 	\item You got sweet pears and lemons: 
 	
 	\lf{ \lam{1}{got}{you, and(\quant{bare}{x}{and(sweet(x), pears(x))}, \quant{bare}{y}{lemon(y)})} }
 \end{description}
 
 \noindent
 but not for sentences in which the modifier has scope over the conjoined structure:
 
 \begin{description}
 	\item You got chocolate eggs and bunnies:
 	
 	\textbf{is}	\lf{ \lam{1}{got}{you, and(\quant{bare}{x}{and(chocolate(x), eggs(x))}, \quant{bare}{y}{bunnies(y)})} }
 	\textbf{should be}	\lf{ \lam{1}{got}{you, \quant{bare}{x}{and(chocolate(x), and(eggs(x), bunnies(x))}}}
 	
 \end{description}

\end{itemize}

\subsubsection*{Open clausal complements}
UD does not distinguish between object control and raising-to-object structures, and so ``I asked you to sit'' and ``I want you to sit'' receive the same UD annotation, despite the fact that ``asked'' semantically takes ``you'' as an argument and ``want'' does not (see \S\ref{sec:UD}).

The converter interprets all open clausal complements as raising-to-object.
 
 \begin{description}
 	\item He wants you to take a nap (\textsc{raising-to-object}):
 	
 	\lf{ \lam{1}{wants}{he, \lam{2}{take}{you, \quant{a}{x}{nap(x)}}} }
 \end{description}
 
 \begin{description}
 	\item Mommy asked you to come (\textsc{object control}):
 	
 	\textbf{is}	\lf{ \lam{1}{asked}{Mommy, \lam{2}{come}{you}} }
 	
 	\textbf{should be}	\lf{ \lam{1}{asked}{Mommy, you, \lam{2}{come}{you}} }
 	
 \end{description}

\subsubsection*{Relative clauses}
As discussed in \S\ref{sec:UD}, UD annotation does not specify the role which the relativized noun takes on in the relative clause. In our UD annotation we subcategorize for subject and object relative clauses, but we do not mark the role of the noun if it is not a core argument. The converter fails on those non-subcategorized relative clauses.

For example, in this case the role that the relativized noun takes in the relative clause is that of an object. The converter therefore produces the correct LF as in here:

 \begin{description}
 	\item all things that you find: 
 	
 	\lf{ \quant{all}{x}{and(things(x), \lam{1}{find}{you, x})} }
 \end{description}

However, in this example, the relativized noun takes the role of a prepositional object in an adjunct landed. Since this is not specified in the UD, the converter will fail on this example, rather produce the correct LF.
 
 \begin{description}
 	\item the spot they landed on:
 	\textbf{should be}	\lf{ \quant{the}{x}{and(spot(x), \lam{1}{and(landed}{they), on\textsubscript{\var{1}}(x)})} }  
 \end{description}

Another difficulty is connected with free relative clauses, in which the head nominal is missing and a relativizer pronoun takes its place, e.g.~``You heard what I said'' in the figure in \S\ref{sec:lfs}. In the LF we treat the wh-word as a determiner, which introduces a variable standing in for the missing nominal.

\subsubsection*{Clauses without overt subject}
All clauses without a subject are assumed to be imperative (this does not include cases of external clausal subject, as in relative clauses, clausal modifiers of nominals, or in raising and control). We are thus assuming an implicit \textit{you} subject and make it explicit in the LF, which can sometimes lead to mistakes.

 \begin{description}
 	\item See you later:
 	
 	\textbf{is}	\lf{ \lamand{1}{see}{you, you}{later}{} }
 	
 	\textbf{should be}	\lf{ \lamand{1}{see}{I, you}{later}{} }
 	
 \end{description}

While it is difficult to precisely quantify the frequency of the constructs described above, 
we do report statistics on the ratio of the utterances for which the converter fails, as well as conduct manual analysis of a sample of produced LFs in order to assess their quality. See \S\ref{sec:experiments}.

\section{Annotating the CHILDES Adam and Hagar Corpora}\label{sec:annotation}

\subsection{The corpora}

\paragraph{Adam.} 
We annotate a total of 17,233 
child-directed utterances from Brown's Adam corpus, covering sessions 1 to 41 and spanning from age 2 years 3 months to 3 years 11 months. 115 utterances which were incomplete (marked by the final token \textit{+...}) were discarded. The corpus contains 107895 tokens.

\paragraph{Hagar.}
We annotate all child-directed utterances in Berman's Hagar corpus, comprising 24,172 utterances in total. The annotated corpus covers 134 sessions (recorded on 115 days, with multiple sessions on some days) from the child's ages of 1 year and 7 months to 3 years and 3 months. 192 incomplete utterances were discarded. The corpus contains 154312 tokens.

We remained faithful to the existing tokenization of the CHILDES corpus, and so any annotation incorporated to Adam or Hagar, was incorporated into the new scheme. 
There have been a number of exceptions to this rule:

\begin{enumerate}
\item
Compounds (tokens that included an underscore $\_$ in them) in the original CHILDES corpus, were split to two tokens, in accordance with the UD guidelines.
\item
Correction of words that had errors: some words (around 100 unique words) had errors, such as ya\#hig\={\i}d instead of yag\={\i}d. These were corrected by replacing the problematic words with the correct ones.
\item 
Splitting possessive pronouns in Hebrew from the main stem: possessive pronouns in Hebrew are generally clitics. In accordance with UD guidelines and the existing annotation for Hebrew, we split the clitics from the main stem.
\end{enumerate}

In addition, incomplete sentences were discarded as well. We have also discarded sentences that contain the tokens ‘xxx’, ‘yyy’ or ‘www’ (indicating unidentifiable material, such as unintelligible words).

\subsection{Annotator training}

The Hagar treebank was annotated by three native speakers of Hebrew with a BA in linguistics. 
The majority of the Adam treebank (13709 utterances) was annotated by a single annotator -- a native English speaker with a BA in linguistics. 
The rest of the corpus (4404 utterances) was annotated by two of the Hebrew annotators, both highly proficient in English. Before annotating the treebank, our annotators received extensive training, which consisted of 1)~a tutorial from a senior member of the team, 2)~reading through the Universal POS tags and English UD guidelines,\footnote{\url{http://universaldependencies.org/}} and 3)~annotating a subset of about 100~sentences from the CHILDES corpus, and discussing issues that came up in the annotation. The Hebrew annotators annotated a training batch of sentences in both languages. 
While working through the training sentences, our annotators met several times with members of the team to seek advice and compare annotations. Upon satisfactory completion of the training sentences, the treebank annotation began.

\begin{figure}
\centering
\includegraphics[width=\textwidth, trim ={6pt 0 6pt 0}]{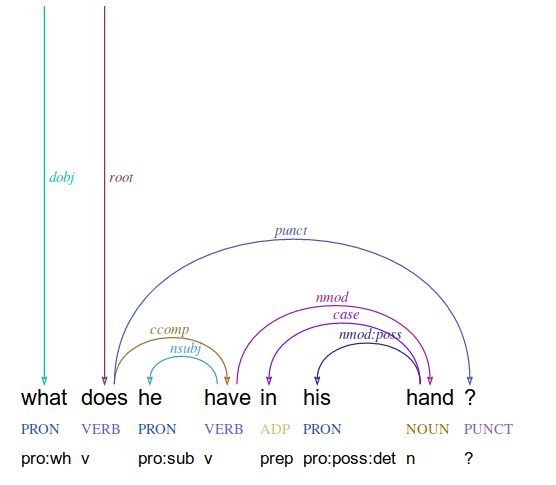}
\caption{A screenshot from the Arborator annotation interface, displaying an automatically converted UD parse, which was later hand-corrected.\label{fig:arborator}}
\end{figure}

\subsection{Annotation Procedure}\label{sec:procedure}

Annotation was carried out using the web-based annotation tool Arborator\footnote{\url{https://github.com/Arborator}} \citep{Gerdes:13}. 
Arborator uses a simple mouse-based graphical interface with movable arrows and drop-down menus to create labeled dependency trees.
In order to expedite the annotation process, we leveraged existing POS tags and dependency trees over the utterances, which were automatically
parsed using the transition-based parser of \citet{Saga:10}, 
and converted to UD through a method based on simple tree regular expressions, using the DepEdit tool\footnote{\url{https://gucorpling.org/depedit/}} \citep{depedit}.
The annotator’s task was then to hand-correct the dependency relations and POS tags as appropriate. 
The code for preprocessing the data and for converting CHILDES dependencies to approximate UDs is freely available online.

Figure~\ref{fig:arborator} presents a pre-annotated sentence given to our annotators through the Arborator interface.
Annotations that the annotators were unsure of were marked as problematic in the annotation tool. Hard cases were extracted and discussed among the members of the team.

\section{Statistics and Evaluation}\label{sec:experiments}

In order to evaluate the self-consistency of the compiled corpora we first measure the agreement between the annotators.
For this purpose, each annotator was assigned a longitudinally contiguous sample of 500 utterances in each of the languages they worked on. 
The starting point of the annotation was the initial converted parser output (see \S\ref{sec:procedure}).

For both Adam and Hagar, we find fairly high agreement scores comparable with those reported in the 
literature for English dependency annotation \citep{berzak2016anchoring},
and somewhat higher than the ones reported for low resource languages \citep{dirix2017universal,nguyen2018bktreebank}.
We obtain a pairwise labeled attachment score (LAS) of 89.9\% on Adam and an unlabeled score (UAS) of 95.0\%, averaging over the three annotators. 
About 0.4\% in the LAS agreement in English is lost due to passive constructions occasionally not marked as such by the Hebrew annotators, possibly due to Hebrew UD not using the passive subject ({\it nsubjpass}) relation.
Average pairwise agreement on Hebrew is 86.7\% LAS and 92.2\% UAS.
While using them facilitates the annotation process, we find that the converted parser outputs
are of fairly low quality: about 40\% of the edges are altered relative to the converted parser output in English, and about 30\% 
of the edges in Hebrew.

Next, we evaluate the UD-to-LF conversion procedure. In terms of coverage, it achieves an 80\% conversion rate on the English corpus and 72.7\% on the Hebrew corpus. We further evaluate the quality by manually evaluating the LFs of a sample of 100 utterances in English and 100 in Hebrew. We find that 82\% of the English LFs in both English and Hebrew are correct.
The LFs we judge to be incorrect generally exhibit at least one of the problems discussed in \S\ref{sec:limitations}.

Table~\ref{pars_stats_english_hebrew} presents statistics of the corpora, including the frequency per token of dependency labels in the full UD annotated corpus as well as in the portion of the corpus which was successfully converted to LF. It should be noted that an occurrence of a dependency type is counted as not converted if the sentence which contains it is not converted. It does not necessarily mean that this particular dependency was the source of the problem. Therefore the conversion rate of a dependency is only a noisy measure of how difficult a given construction is for the converter.


\begin{table}

	\small
	\centering
	{ 
		\resizebox*{!}{.95\textheight}
		{\begin{tabular}{ l >{\raggedleft}p{2.5em}  c | l  >{\raggedleft}p{2.5em}  c}
\multicolumn{3}{c}{\textbf{Adam}} & \multicolumn{3}{c}{\textbf{Hagar}}\\
\midrule
\textbf{dep label} & \multicolumn{1}{c}{\textbf{count}} & \textbf{\% converted} & \textbf{dep label} & \multicolumn{1}{c}{\textbf{count}} & \textbf{\% converted}\\
\midrule
list & 56 & 96 & csubjpass & 1 & 100\hphantom{0} \\
xcomp:promoted & 40 & 82 & vocative & 1863 & 71\\
quant & 11 & 82 & ccomp:promoted & 14 & 71\\
nmod:poss & 1567 & 78 & compound:svc & 128 & 70\\
vocative & 805 & 77 & root & 18783 & 68\\
aux & 6048 & 76 & acl:promoted & 3 & 67\\
parataxis:repeat & 48 & 75 & cop & 580 & 66\\
root & 14724 & 74 & dislocated & 165 & 65\\
det & 6278 & 74 & dobj & 6171 & 64\\
punct & 14858 & 74 & parataxis:repeat & 960 & 64\\
nsubj & 12816 & 74 & nsubj & 14401 & 62\\
dep & 23 & 74 & punct & 29104 & 61\\
discourse & 2412 & 74 & nmod:smixut & 615 & 60\\
dobj & 6774 & 74 & nmod & 8160 & 59\\
amod & 1138 & 74 & name & 132 & 59\\
dislocated & 11 & 73 & det & 9439 & 59\\
nummod & 259 & 73 & amod & 2185 & 58\\
case & 4434 & 72 & case & 12990 & 57\\
cop & 3122 & 72 & compound:prt & 21 & 57\\
neg & 2306 & 72 & ccomp & 1309 & 57\\
nmod & 3617 & 71 & acl:relcl:subj & 151 & 56\\
compound:svc & 59 & 71 & compound:smixut & 32 & 56\\
name & 183 & 70 & acl:relcl:obj & 228 & 54\\
acl:relcl:obj & 153 & 70 & nmod:poss & 958 & 54\\
dobj:promoted & 128 & 70 & advmod & 6762 & 54\\
advmod & 5194 & 69 & discourse & 3892 & 52\\
compound:prt & 288 & 68 & nummod & 199 & 50\\
compound & 1003 & 68 & fixed & 8 & 50\\
iobj & 262 & 67 & advcl:promoted & 8 & 50\\
acl:promoted & 3 & 67 & root:promoted & 359 & 50\\
remnant & 12 & 67 & mark & 2243 & 48\\
goeswith & 42 & 67 & cc & 3470 & 46\\
nmod:promoted & 39 & 64 & parataxis & 3876 & 46\\
acl:relcl:subj & 114 & 63 & xcomp & 1848 & 45\\
ccomp & 1140 & 63 & list & 43 & 44\\
csubj & 8 & 63 & goeswith & 44 & 43\\
nsubj:promoted & 18 & 61 & nsubjpass & 48 & 42\\
parataxis & 459 & 59 & reparandum & 257 & 40\\
mark & 1973 & 59 & remnant & 32 & 38\\
advcl & 568 & 58 & advcl & 573 & 35\\
xcomp & 1405 & 57 & mwe & 831 & 35\\
reparandum & 42 & 52 & nmod:promoted & 127 & 35\\
nsubjpass & 35 & 51 & csubj & 93 & 33\\
auxpass & 32 & 50 & det:predet & 157 & 33\\
comp & 2 & 50 & conj & 1911 & 30\\
nmod:npmod & 26 & 46 & xcomp:promoted & 7 & 29\\
cc & 881 & 44 & compound & 492 & 28\\
det:predet & 64 & 42 & expl & 69 & 28\\
mwe & 85 & 40 & aux & 170 & 19\\
nmod:tmod & 128 & 39 & nmod:tmod & 158 & 18\\
expl & 211 & 38 & acl & 42 & 17\\
ccomp:promoted & 32 & 34 & dep & 516 & 12\\
acl & 72 & 33 & nsubj:promoted & 120 & 10\\
conj & 675 & 32 & dobj:promoted & 135 & \hphantom{0}7\\
advcl:promoted & 7 & 29 & appos & 267 & \hphantom{0}4\\
root:promoted & 107 & 21 & case:gen & 36 & \hphantom{0}3\\
appos & 26 & \hphantom{0}8 & acl:relcl & 85 & \hphantom{0}1\\
csubjpass & 1 & \hphantom{0}0 & csubj:promoted & 2 & \hphantom{0}0\\
acl:relcl & 100 & \hphantom{0}0 & acl:relcl:subj:promoted & 1 & \hphantom{0}0\\
		\end{tabular}}}
		\caption{Dependency label counts and proportion of dependencies which were successfully converted to LFs for the Adam corpus (left) and Hagar corpus (right). Ordered by \% of occurrences converted.}
	\label{pars_stats_english_hebrew}
\end{table}

\section{Corpus Analyses}\label{sec:analyses}

\aumod{This section provides some initial analyses of both the syntactic and semantic aspects of our corpora. While simple, we hope these analyses, together with the modelling study in \S\ref{sec:simulations}, will provide inspiration to other researchers regarding some of the questions that can be examined using these resources.}
\oamod{Results here are mostly intended to demonstrate the potential utility of the proposed dataset, and should be interpreted with caution, taking into account the inter-annotator agreement (see \S\ref{sec:experiments}).
The analysis is based on the dependency structures, rather than LFs.
The reason for doing so is that UD is annotated over other, non-CDS corpora in both English and Hebrew, which allows comparing the statistics of the compiled corpora to those of existing ones.}

\subsection{Analyses of syntactic dependencies}

\aumod{
This section highlights some of the benefits of using the Universal Dependencies scheme. In particular, since this scheme is also used for adult-directed language, we can quantify some of the differences between our child-directed corpora and existing text corpora (\S\ref{sec:adult_ud_compare}). Perhaps of more interest to language acquisition researchers, the cross-linguistic consistency of the UD scheme also permits direct comparisons between the input to the child in different languages, as we demonstrate in \S\ref{sec:cds_ud_compare}. Longitudinal analyses are also possible, as shown in \S\ref{sec:ud_longitudinal}. 

While our analyses are very simple frequency comparisons, other researchers might be interested in more subtle analyses, for example using the UD annotations to search for particular constructions of interest in one or more languages, to analyze these in more detail.
}

\subsubsection{Comparison to general corpora of English and Hebrew}
\label{sec:adult_ud_compare}

The dependency statistics of our CHILDES corpora can be compared to those of general treebanks of written English and Hebrew, English Web Treebank \citep{silveira2014gold} and Hebrew Dependency Treebank \citep[HDT;][]{tsarfaty2013unified,mcdonald2013universal} respectively. Statistics are based on the entire corpora, ignoring the split into training, development and test sets. We focus our study comparison on the dependency annotation (rather than the LFs), as dependency structures decompose straightforwardly to atomic elements that can be counted and compared, and thus lend themselves more easily to statistical analysis.

\begin{figure}
\centering
\begin{subfigure}{\textwidth}
  \centering
     \includegraphics[width=\linewidth, scale=0.8, trim={0 0 1cm 0}]{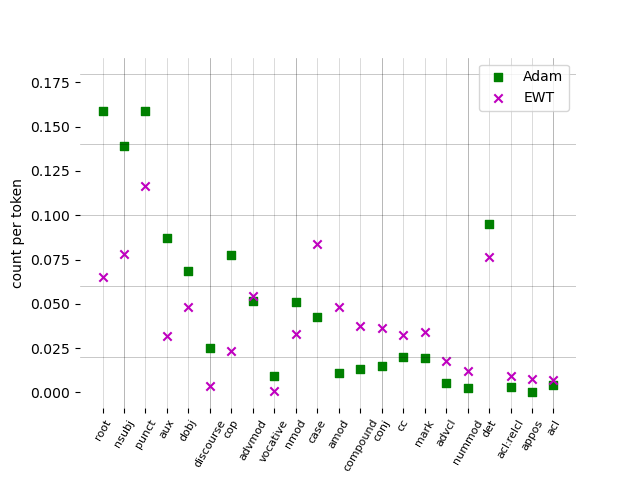}
  \caption{English corpora (Adam and EWT)}\label{fig:english_corpus_comparison}
\end{subfigure}
\begin{subfigure}{\textwidth}
  \centering
  \includegraphics[width=\linewidth, trim={0 0 1cm 0}]{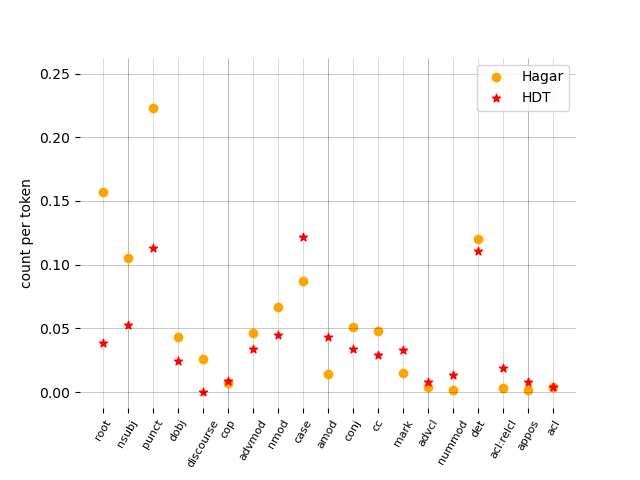}
  \caption{Hebrew corpora (Hagar and HDT)}\label{fig:hebrew_corpus_comparison}
\end{subfigure}
\caption{Comparison of dependency type prevalence in child-directed speech and standard UD corpora of the same language. The plots show only dependencies with a difference in count per token of $>$ 0.005 \aumod{between each CDS corpus and its paired general text corpus.
In each plot, dependencies are sorted according to the size of this difference: starting from the left are the dependencies with greater prevalence in CDS (sorted from larger to smaller differences with general text), followed by those with greater prevalence in general text (again, sorted from larger to smaller differences with CDS).}
}
\label{fig:corpus_comparison}
\end{figure}

\paragraph{Adam.} 
As can be seen in Figure~\ref{fig:english_corpus_comparison}, 
not many dependency types are more frequent in child-directed language than in general English. The Adam corpus exhibits a higher prevalence of discourse phenomena and direct address to the interlocutor (\textit{vocative}), which is explained by virtue of it being a corpus of conversational spoken language.

The higher frequency of basic relation types (\textit{root, punct, nsubj, dobj}, and \textit{aux}) is a result of the sentences being shorter than in the EWT corpus (a mean of 5.9 tokens per sentence as compared to 15). We also note that negation is more frequent in our corpus than in general English, and so is adverbial modification. The latter is perhaps attributable to a large number of questions about ``why'' and ``how'' in our corpus.
Structures markedly more common in general English include adjectival modification, conjunction, compounding, prepositional phrases, clausal modifiers and passive voice. A slight difference is observed in the frequency of determiner use, possibly reflecting the fact that in the child-directed corpus we find many examples of naming things or affirming the child's utterance in which bare nominals are used, e.g.~``Yes, scout'', ``Ice for boys and girls''.

\paragraph{Hagar.}
Comparing the Hagar corpus to HDT (Figure~\ref{fig:hebrew_corpus_comparison}) we again observe higher frequency of the core dependencies in child-directed language because of the difference in average utterance length (an average of 6.4 tokens per sentence in the Hagar corpus and 19 tokens in HDT). The more discursive nature of the Hagar corpus is reflected in the higher prevalence of the \textit{parataxis} and \textit{discourse} relations.  As in the case of English, negation and adverbial modification are slightly less frequent in general Hebrew. In contrast to English, however, the  \textit{aux} relation is more common in HDT than in the Hagar corpus. In Hebrew UD auxiliaries often express modality or aspect, which might characterize news text (source of HDT data) more than child-directed language.\footnote{In the Adam corpus the most common auxiliaries are forms of \textit{do}, \textit{be}, and \textit{can}, reflecting the high frequencies of do-support, copular constructions, asking the child to do something or answering questions about things being possible or allowed.}

Similarly to English, general Hebrew displays noticeably higher frequencies of adjectival modification, conjunction, compounding, prepositional phrases, and clausal modifiers, but also possessives and indirect objects. The difference in \textit{iobj} frequency might be attributable to the HDT corpus assuming different annotation guidelines and using the \textit{iobj} label where we use \textit{nmod}.\footnote{In English, \textit{iobj} applies to the first object in the double object construction (e.g.~\textit{Give \textbf{me} the book}). In Hebrew, this construction is very uncommon, and ditransitives in English are often translated to a dative PP  (like \textit{Give the book \textbf{to me}}).} The frequency of determiner use is much higher in HDT, which may be explained by the lower frequency of  \textit{amod} and \textit{nmod} in Hagar. These two dependency relations are the most common edge labels of the determiner heads in HDT (over 60\% of the total number of such edges).

\subsubsection{Comparison of Adam and Hagar corpora}
\label{sec:cds_ud_compare}

\begin{figure}
\centering
\includegraphics[width=\linewidth,trim={0 0 1cm 0}]{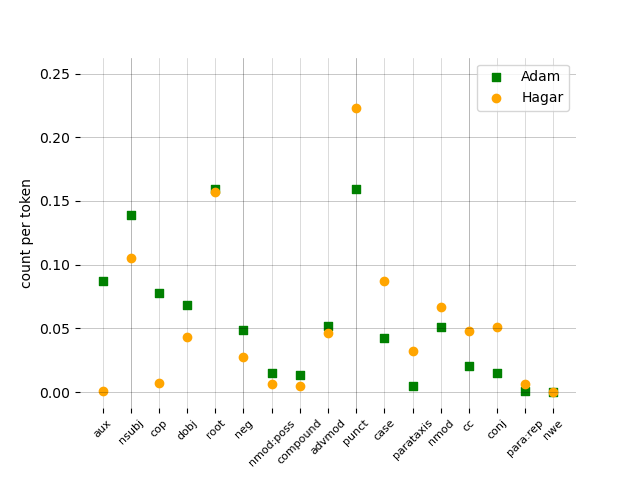}
\caption{\aumod{Comparison of dependency type prevalence between the English and Hebrew CDS corpora. The plots show only dependencies with a difference in count per token of $>$ 0.005, and are sorted according to the size of this difference: starting from the left are the dependencies with greater prevalence in the Adam corpus (sorted from larger to smaller differences with Hagar), followed by those with greater prevalence in the Hagar corpus (again, sorted from larger to smaller differences with Adam).}}
\label{fig:adam_hagar_comparison}
\end{figure}

Figure~\ref{fig:adam_hagar_comparison} compares the two CHILDES corpora. There are not many notable differences between the frequency of occurrences of particular dependency relations between the English and Hebrew corpora. Sentences in the Adam corpus are on average shorter (5.9 tokens per utterance as compared to 6.4), which is reflected in the higher frequency of \textit{root}, \textit{nsubj}, and \textit{dobj} relations in Figure~\ref{fig:adam_hagar_comparison}. The difference in \textit{nsubj} is likely also related to Hebrew being a pro-drop language. Other differences also reflect diverging properties of the two languages: prevalence of \textit{cop} in English is higher, because Hebrew lacks an overt copula; prevalence of \textit{aux} is higher in English, since tense, which accounts for many of the \textit{aux} instances in English, is encoded morphologically in Hebrew; prevalence of \textit{case} and \textit{nmod} in Hebrew is higher likely because of indirect objects being expressed using case markers. 

Other observed differences, like more negation and possessives in English or more adjectives, conjunctions, and parataxis in Hebrew, might be idiosyncratic to the speakers.
Other differences may be due to different transcription conventions. For instance, the Hebrew corpus contains markedly more commas.

\subsubsection{Longitudinal analysis of syntactic dependencies}\label{sec:ud_longitudinal}

Taking advantage of our chronologically ordered data we inspect the changes in frequency of use of particular dependency labels over time. For each dependency and each session, we calculate the proportion of sentences which include that dependency. We check for the existence of longitudinal trends by examining whether the child's age is a significant predictor, in a linear regression model, of the frequency of each dependency. Below we discuss dependencies which exhibit a trend with p $ < $ .01.

\begin{figure}
\centering
\includegraphics[width=0.5\linewidth]{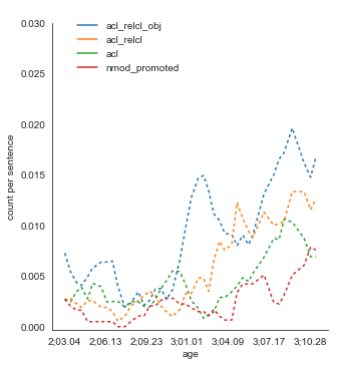}
\caption{Dependencies displaying an upward longitudinal trend in frequency in the Adam corpus. Frequencies are smoothed over 5 sessions.}
\label{fig:adam_trends}
\end{figure}

\begin{figure}
\centering
\begin{subfigure}{0.48\textwidth}
  \centering
  \includegraphics[width=\linewidth]{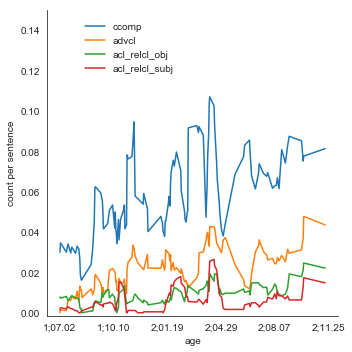}
  \caption{clausal complements, adverbial clauses, relative clauses}
\end{subfigure}
\begin{subfigure}{0.48\textwidth}
  \centering
  \includegraphics[width=\linewidth]{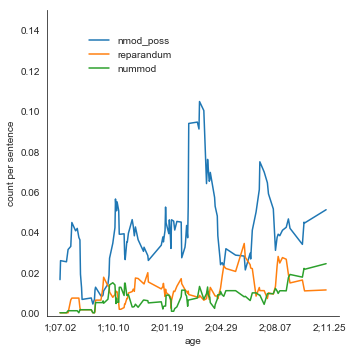}
  \caption{possessive, disfluencies, and numerical modifiers of nouns}
\end{subfigure}
\begin{subfigure}{0.48\textwidth}
  \centering
  \includegraphics[width=\linewidth]{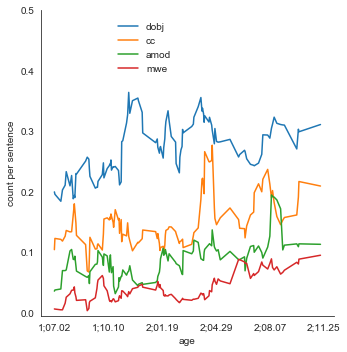}
  \caption{direct object, conjunction, adjectives, multiword expressions}
\end{subfigure}
\begin{subfigure}{0.48\textwidth}
  \centering
  \includegraphics[width=\linewidth]{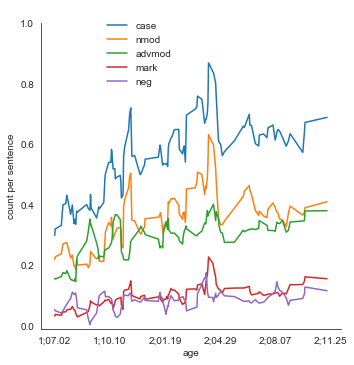}
  \caption{prepositional phrases, adverbs, subordinate clauses, negation}
\end{subfigure}
\caption{Dependencies displaying an upward longitudinal trend in frequency in the Hagar corpus. Frequencies are smoothed over 5 sessions. (The grouping of dependencies is not meaningful but merely increases legibility)}
\label{fig:hagar_trends}
\end{figure}

In the Adam corpus, we find a significant increase in the use of the following constructions as the child gets older: adjectival clauses, object and ``other'' (i.e., not subject or object) relative clauses, and  ellipsis  affecting  nouns  in  prepositional  phrases (Figure~\ref{fig:adam_trends}). In the Hagar corpus longitudinal changes are much more widespread (Figure~\ref{fig:hagar_trends}). The point of commonality is relative clauses--in the case of the Hagar corpus there are upward trends for subject and object relatives. The following constructions also significantly increase in use with time: adverbial clauses, adjectives, numerical and possessive modifiers, multiword expressions, disfluencies (reparandum), transitive verbs (direct object), conjunction, adverbs, subordinate clauses (mark), clausal complements, negation, and prepositional phrases, which in our annotation include all indirect objects.

\subsection{Longitudinal analysis of semantic complexity}

\aumod{As well as syntactic analyses, our corpora provide meaning representations, which allow additional types of research questions. Again, we provide just a simple proof of concept here, investigating whether the semantic complexity of the adults' utterances increases as the child gets older.} 
\oamod{Future work may wish to conduct other types of analyses that are not explicit in the UD syntax but are exposed by the LFs, such as statistics on the valency of different predicates, or the scope of quantifiers.}

\subsubsection{Semantic complexity measures}

In the context of this corpus analysis, we propose a very constrained definition of semantic complexity. We consider complexity in the sense of structural complexity of the predicate-argument relationships in the utterance -- the depth of nesting and the number of predicates, arguments and modifiers.

The most pertinent question when it comes to the longitudinal analysis of the semantic complexity of CDS is whether the adult utterances express increasingly complex meanings as the child gets older. There are many axes on which complexity can increase -- concepts being more abstract, referents of expressions being less contextually obvious, language being more metaphorical, etc. 
Our newly available data creates the opportunity to study this question in terms of sentential predicate-argument structures. That is, the question we answer in this analysis is: does the predicate-argument structure of the CDS grow more complex as the child grows older?

 \begin{figure}
 \begin{minipage}{0.64\textwidth}     
 The full LF for the sentence \textit{``What happened to your finger?''} is shown below:
\begin{description}
 	\item
  	\lf{ \lamwh{a}{\lamand{1}{happened}{a}{to}{(\quant{your}{x}{finger(x)})}}}
 \end{description}
From it, we extract the following five sub-expressions (corresponding to the dashed boxes in the tree):
\begin{itemize}
 	\item \lf{ finger(x) }
    \item \lf{ \quant{your}{x}{finger(x)} } 
    \item \lf{ to\textsubscript{\var{1}}(\quant{your}{x}{finger(x)}) } 
  	\item \lf{ happened\textsubscript{\var{1}}(a) }
  	\item \lf{ \lamand{1}{happened}{a}{to}{(\quant{your}{x}{finger(x)})} }
 \end{itemize}
 \end{minipage}
\begin{minipage}{0.35\textwidth}     
\includegraphics{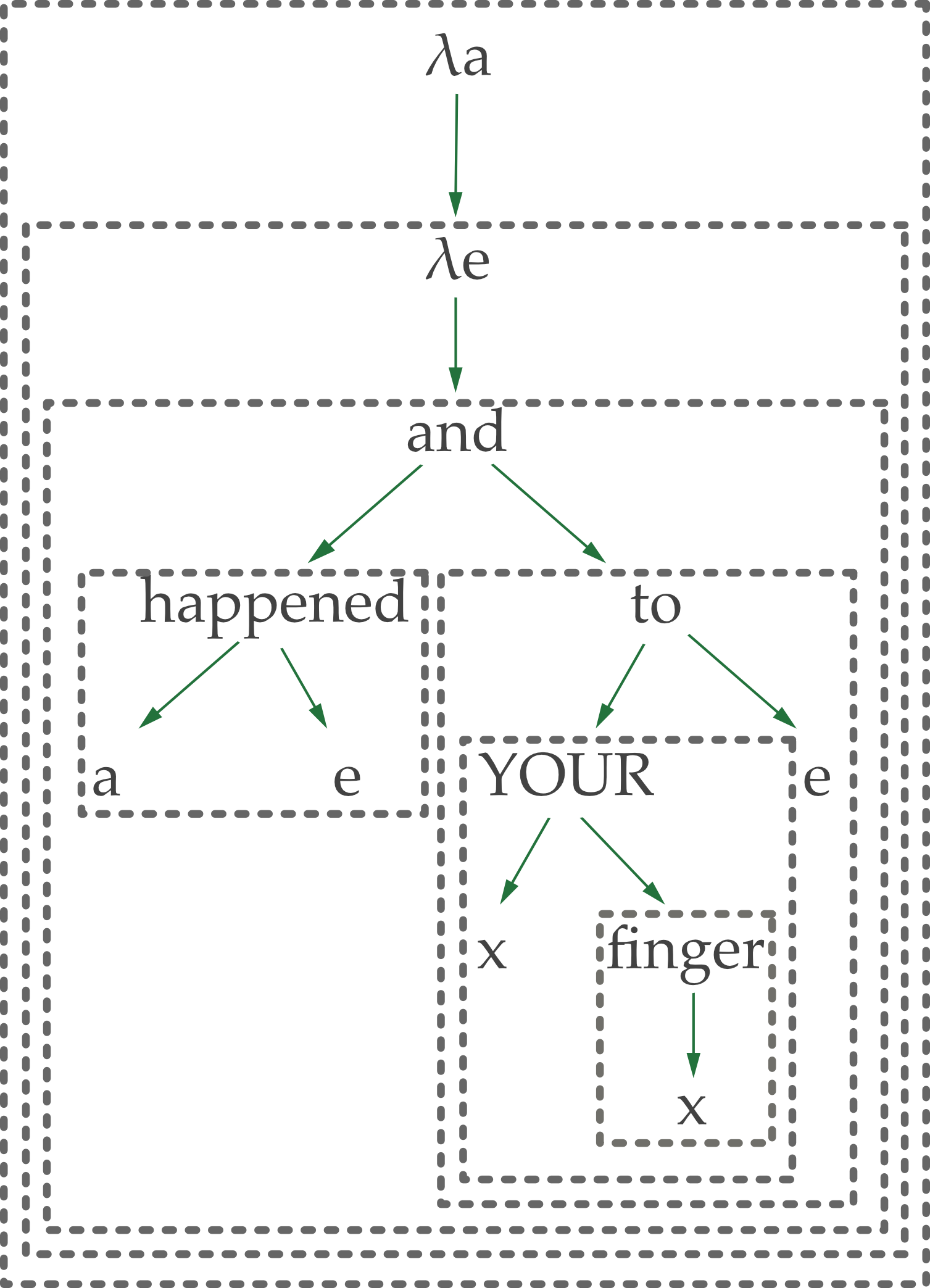}
  \end{minipage}

  \caption{Extracting sub-expressions from an example LF. If one views the LF as a tree in which variable bindings, predicates, and variables are nodes, then sub-expressions correspond to sub-trees of that tree (indicated by dashed boxes). The number of sub-expressions reflects both the branching factor and nesting level of the tree. 
  }
  \label{fig:subexpressions}
\end{figure}

One way to approach the issue is to count the number of sub-expressions in the LF. 
For example, from the LF of the sentence ``What happened to your finger?'', \aumod{we obtain five sub-expressions, as illustrated in Figure~\ref{fig:subexpressions}.}

\begin{figure}
\centering
\begin{subfigure}{0.5\linewidth}
  \centering
  \includegraphics[width=\linewidth]{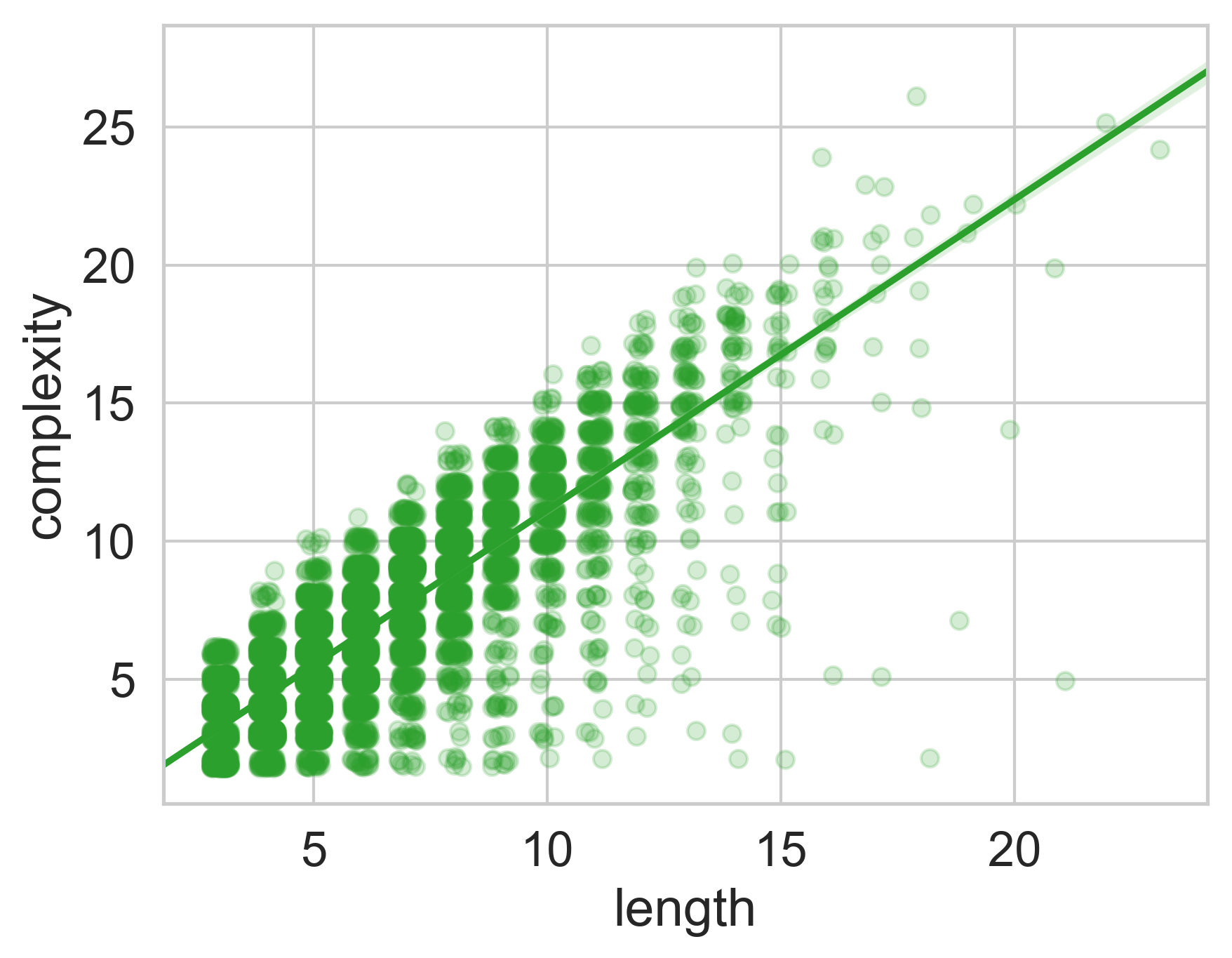}
  \caption{Adam}
  \label{fig:c_l_adam}
\end{subfigure}
\begin{subfigure}{0.5\linewidth}
  \centering
  \includegraphics[width=\linewidth]{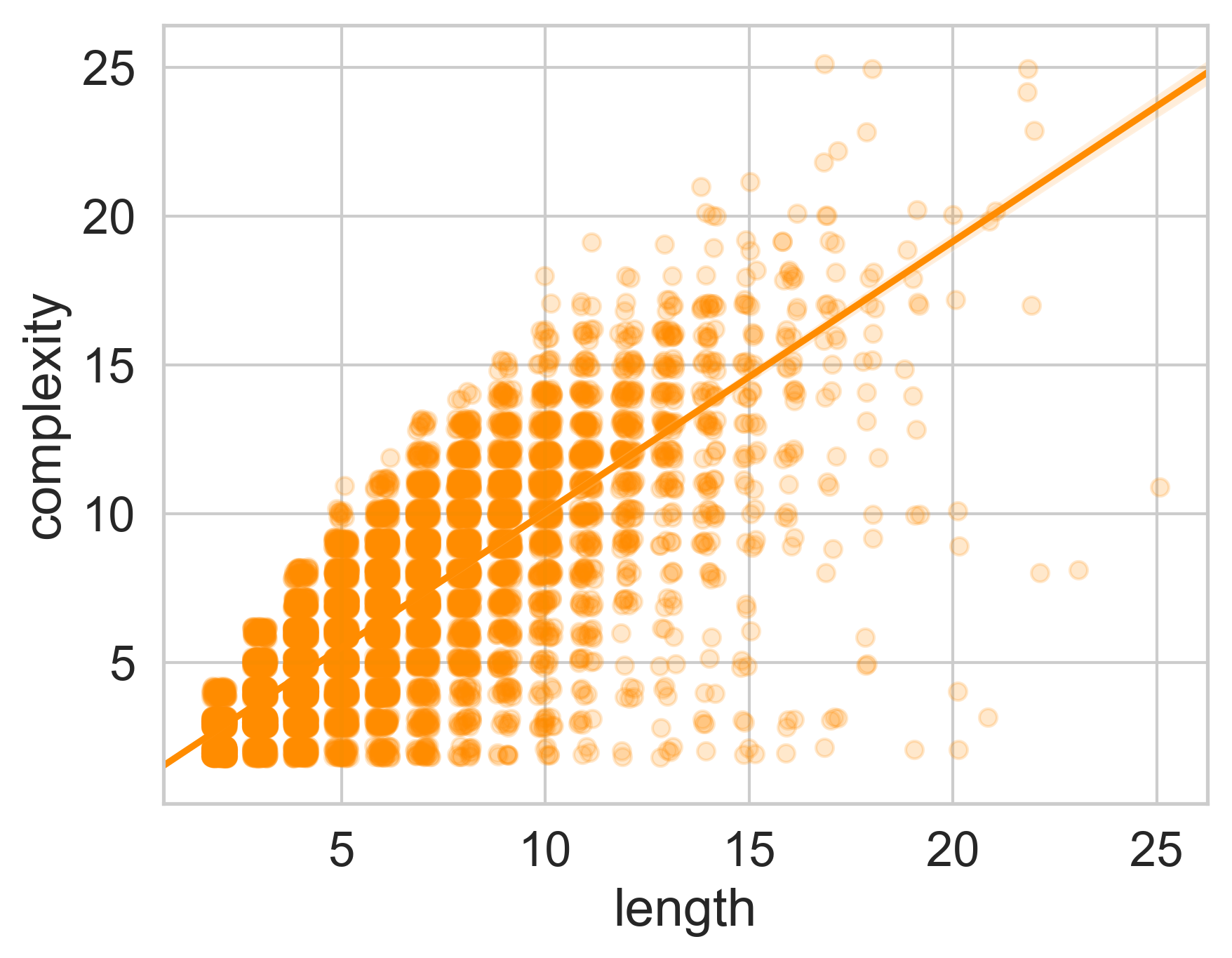}
  \caption{Hagar}
  \label{fig:c_l_hagar}
\end{subfigure}
\begin{subfigure}{0.5\linewidth}
  \centering
  \includegraphics[width=\linewidth]{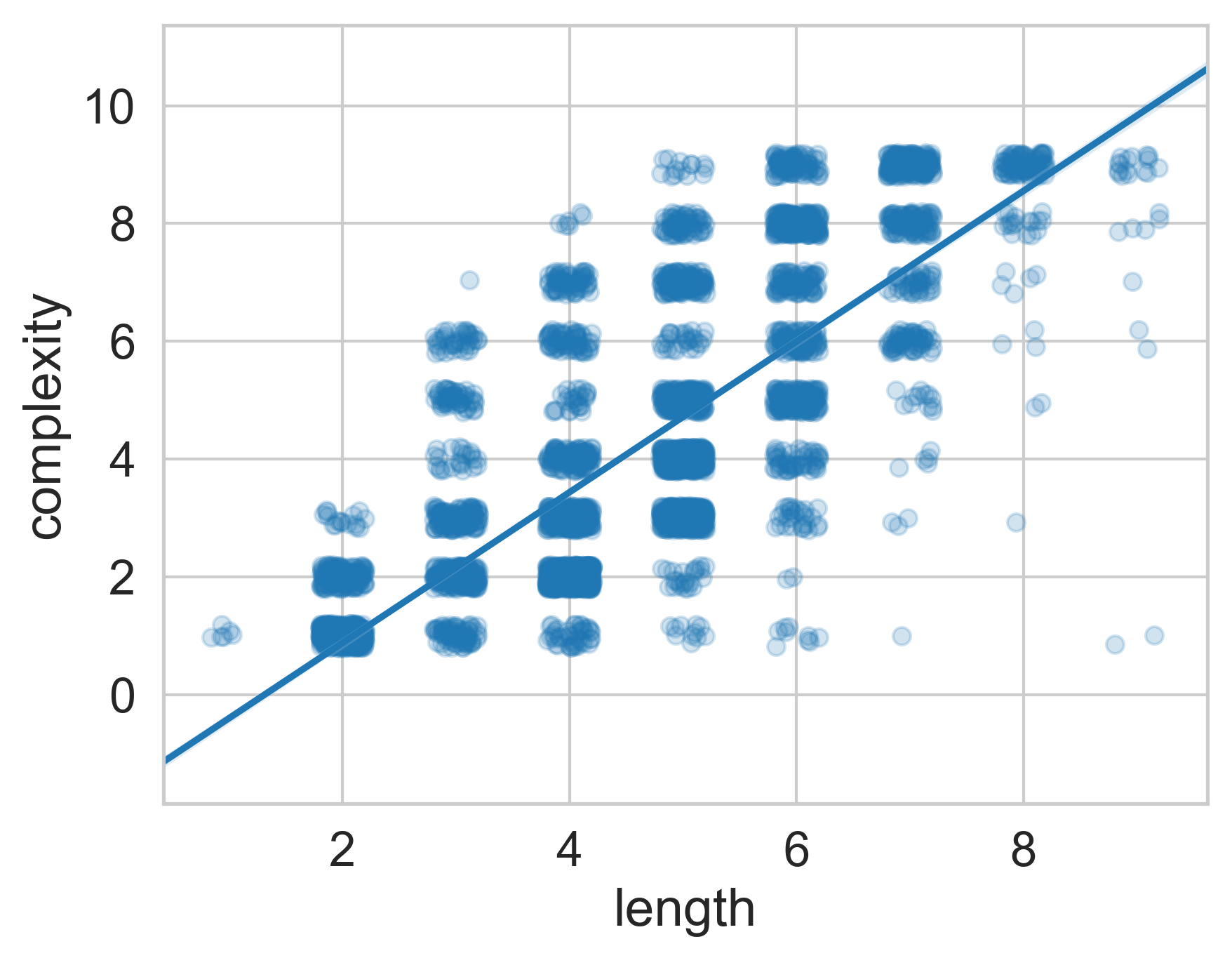}
  \caption{Eve}
  \label{fig:c_l_eve}
\end{subfigure}
\caption{Relationship between LF complexity (number of sub-expressions) and utterance length (number of tokens). Each point represents an individual utterance in the corpus. Solid lines illustrate the linear regression line and the shaded region around the lines the 95\% confidence interval for that regression (very tight in all 3 graphs).}
\label{fig:c_l}
\end{figure}

It is expected that the number of sub-expressions will correlate strongly with the number of tokens in the utterance. For both corpora it is indeed the case, as can be seen in Figure~\ref{fig:c_l} with the Pearson's coefficient of $r=0.74$ for Adam and $r=0.76$ for Hagar. Even though the correlation is strong, we can also observe a relatively wide spread of complexities for any given value of length. The coefficient of determination in OLS regression shows that utterance length accounts for 54.1\% of variation in complexity in Adam and 58.1\% in Hagar. This indicates that our complexity measure captures information beyond just the number of tokens in an utterance. 

To illustrate the improvement of our automated approach of LF generation over a more restricted dataset, and in particular to highlight the usefulness of the relatively high coverage of syntactic constructions in our transducer, we also analyse the semantic complexity of Brown's Eve dataset \citep{Brow:73}. \oamod{To this end we use the transduced LFs by \citet{Abend:17}, which were created semi-automatically from the morphosyntactic annotation of \citet{Saga:10}, and filtered to only include utterances of length up to 10, due to the limitations of their conversion method.
Results (Figure~\ref{fig:c_l}) present a seemingly even stronger correlation and less variation on the Eve dataset than for Adam and Hagar. In the case of the Eve corpus, utterance length accounts for 66.2\% of the variation in complexity.  }

\subsubsection{Longitudinal analysis}
\begin{figure}
\centering
\includegraphics[width=0.95\linewidth]{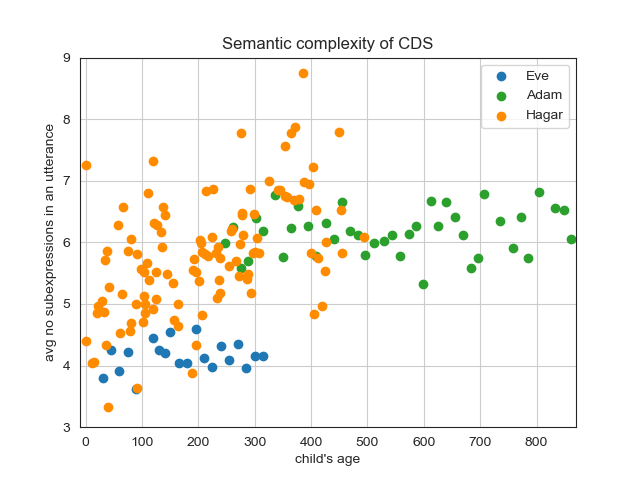}
\caption{The average complexity of child-directed utterances in a session plotted against the child's age in days.}
\label{fig:LF_complexity_all}
\end{figure}

\paragraph{Does CDS complexity change with the child's age?}
Figure~\ref{fig:LF_complexity_all} shows the distribution of semantic complexity (averaged over all utterances in a session) relative to the child's age. 
While the complexity in the Adam corpus remains relatively stable over time, there might be an upwards longitudinal trend in Hagar. Pearson's coefficient confirms a weak correlation between average complexity and child's age in Hagar ($r=.35$, $p<0.001$) and no correlation in Adam ($r=.11$, $p=.5$). Looking further into the Hagar corpus, the OLS regression's coefficient of determination indicates that the child's age accounts for 11\% of the variation in complexity \aumod{beyond that explained by utterance length alone}. Interestingly, Figure~\ref{fig:LF_complexity_all} also suggests that the LFs in the Eve corpus might not adequately reflect the longitudinal changes in CDS utterance complexity. The Hagar corpus presents an increase in semantic complexity at the age range covered by Eve, and we therefore would have expected to see a similar one in Eve. 
\oamod{The fact that such a trend is not observed
is probably due to some limitation of the Eve LFs, which were formed by a
different extraction method to ours (see \S\ref{sec:analyses}). We would expect an increase in meaning
complexity in line with the child's cognitive development over this age range.
The fact that our LFs show an increase in complexity may suggest that they
capture the relevant semantic information in the text, and if so, this is evidence for the
superiority of our method over the method used for compiling the Eve LFs dataset, which does not reflect such a trend.
However, our results do not allow us to decide whether this is the
case or not, and there may be individual or cultural differences
across our data.
}

\section{LF annotated corpora as data for acquisition simulations}
\label{sec:simulations}

This section presents a set of preliminary experiments that demonstrate how the presented
corpora may be used for simulations of the learning dynamics that resemble child language acquisition processes in children.
The cross-linguistic consistency of the scheme allows us to evaluate the cross-linguistic applicability of the model, which is essential for establishing the validity of an acquisition model.

\subsection{The Learning Model: An Outline}

We adapt the language acquisition model of \citet{Abend:17} to the proposed LFs. The model is a computational implementation of the semantic bootstrapping hypothesis \citep{Bowerman:73b,Pink:79}, whose goal is to generalize from input pairs of observed utterances and inferred meanings in
order to interpret new utterances whose meaning is unavailable contextually. Unlike ``parameter-setting'' approaches \citep[e.g.,][]{Yang:02}, we do not assume that the grammar of
natural languages can be described by a finite number of finitely-valued parameters. Instead, the proposed model
searches a structured space of all possible grammars as defined by an established formal theory of the syntax-semantics interface -- CCG.

The proposed model employs Bayesian learning to jointly model (a) learning of the lexicon: the mapping between words (or
generally: any portion of the input string) and portions of the sentential meaning, and (b) syntax learning: the rules
governing the combination of the lexical elements into utterances. By jointly modeling lexical
learning and syntactic acquisition, and assuming that the inferred meanings available to the child are at the level of
utterances rather than individual words, the model provides a working account of how these two aspects of language
can be learned simultaneously in a mutually reinforcing fashion. 

\subsection{Learning of Word Order}

Both English and Hebrew are regarded as languages with Subject-Verb-Object as their basic word order. We report here experiments that show that when experimenting with the learner on the proposed corpora, the probability of SVO indeed increases during learning. 

\oamod{We run the learner on the Adam and Hagar corpus, with their corresponding LFs, and compare our results to those reported by \citet{Abend:17} for the Eve corpus (using length-bounded sentences and using a different, semi-automatic approach for generating the LFs). Experiments are performed without introducing any intentional noise to the training data, and therefore correspond to their ``No Distractors'' setting.}

\begin{figure}
\centering
\begin{subfigure}{.45\textwidth}
  \centering
  \includegraphics[width=1\linewidth]{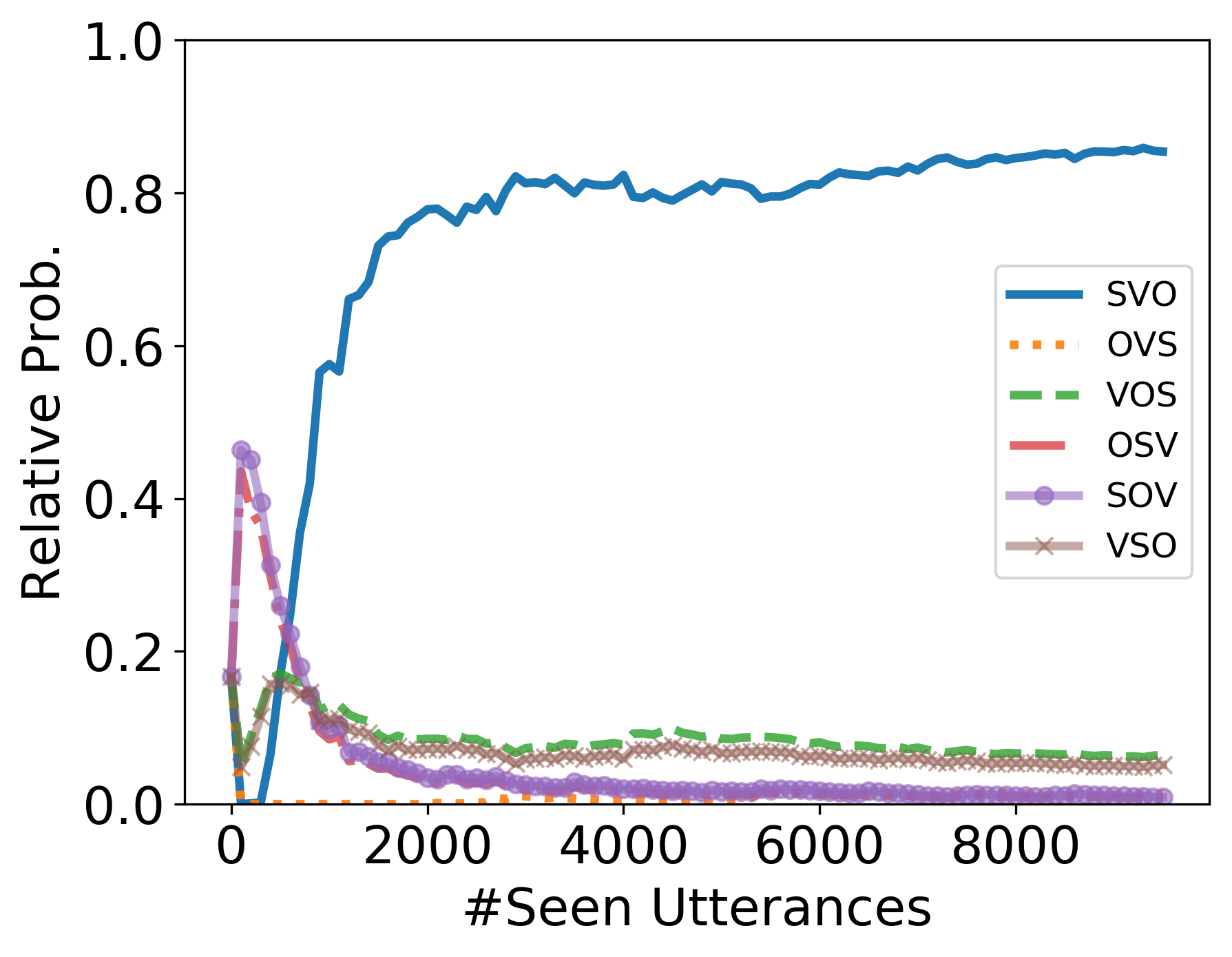}
  \caption{Adam}
  \label{fig:Adam_SVO}
\end{subfigure}%
\begin{subfigure}{.45\textwidth}
  \centering
  \includegraphics[width=1\linewidth]{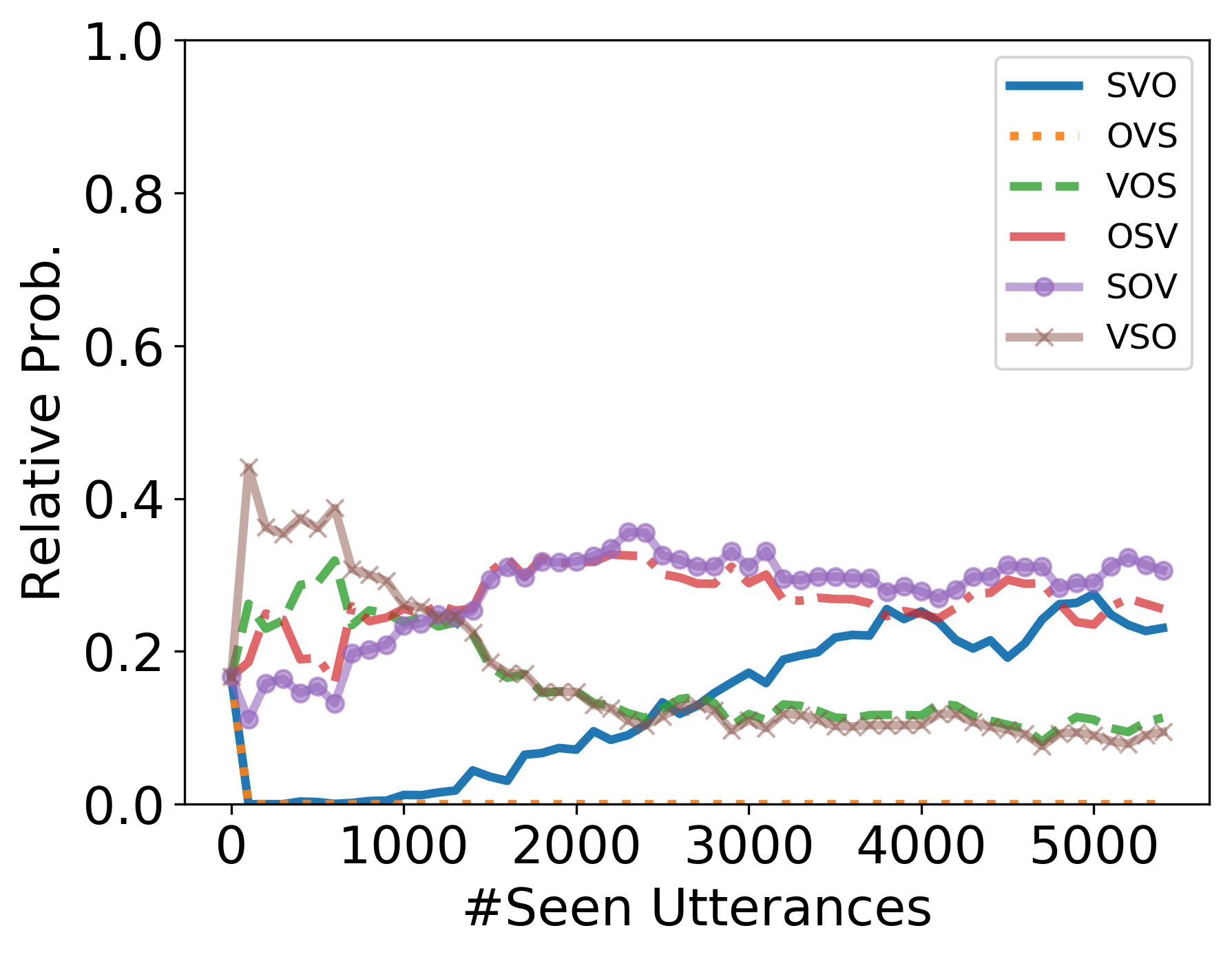}
  \caption{Hagar}
  \label{fig:Hagar_SVO}
\end{subfigure}
\caption{Learning that English and Hebrew are SVO languages. Plots show the relative posterior probability assigned by the model to the six possible categories of transitive verbs.}
\label{fig:SVO}
\end{figure}

\oamod{Figure~\ref{fig:SVO} presents our results.
On Adam the model learns that English transitive sentences are SVO; learning curves are steep, despite the lack of an explicit signal. Comparing the trends to the ones reported on Eve by \citet{Abend:17}, we find that while SVO emerges as the overwhelmingly most probable order in both cases, in the simulation based on the Adam corpus, other orders are considered more probable for around the first 1000 utterances, while with the Eve corpus, SVO overwhelms other hypotheses within the first 100 sentences. It appears that the Eve corpus is too limited in terms of sentence structure variety and complexity to allow for examining the period of acquisition before the basic word order is determined in the mind of the learner.}

In Hebrew, learning is considerably more gradual. In fact, after training on 4000 utterance-LF pairs, the model has managed to demote the (incorrect) VSO, VOS and OSV orders, but remains indecisive as to whether SVO or the verb-final orders are correct. A steady increase, however, is presented in the probability of the correct SVO order. This more gradual trajectory may be due to the more flexible word order presented by Hebrew, as opposed to the relative rigidity of English.

We report on these experiments in order to illustrate the potential usefulness of the corpora and LFs reported here.
A deeper investigation of these and other trends in the acquisition of grammar is needed in order to draw cognitive conclusions from such experiments.

\section{Discussion}\label{sec:discussion}

Having demonstrated the utility of our resource, we consider limitations and opportunities for future extensions.

First, we note that the approach of annotating dependency syntax and automatically transducing it to logical forms is practical but not perfect. We have discussed limitations of the logical forms (\S\ref{sec:limitations}) and estimated the error present in syntactic annotation and conversion to semantics (\S\ref{sec:annotation}). We believe the accuracy is sufficient for examining broad trends (e.g., over the course of acquisition, as in the pilot study in \S\ref{sec:experiments}). But further work on the representations may be required to support research that relies on grounding in a world model, for example.

Second, the style of semantic representation (as rather conventional logical forms in the formal semantics tradition) is suited to some modes of investigation but not all. 
Other semantic representations that exist in broad-coverage corpora might better capture elements such as discourse context \citep{Kamp:93}, lexical semantics \citep{Banarescu:12,pustejovsky1998generative}, or typologically motivated scene structures \citep{abend2013universal}.
Future studies might profit from enriching our data with such representations, building on the LFs that are there to expose syntactically nonlocal semantic dependencies.

Third, the corpus has been designed to facilitate research on the semantic bootstrapping hypothesis. As such, semantic representations are provided for child-directed speech, in order to simulate the meaning that is presumably available to the child in an interaction.
Our focus on child-directed speech is typical of much of the acquisition literature using CHILDES data \citep[{\it inter alia}]{fazl:10,Perfors:11b,huebner-etal-2021-babyberta,aditya2023how} Nevertheless, some lines of research may benefit from syntactic and semantic annotation of child utterances as well. Annotating child language is difficult because it requires interpretation of utterances exhibiting non-mature syntax, and guidelines to support this \citep[cf.~UD annotation of adult learner syntax;][]{berzak2016universal}. We leave this to future work.

Finally, we have investigated two languages in this study as a case study of the considerations needed for cross-linguistic work with our approach. Two languages are, of course, not sufficient to demonstrate that a representation is ``universal'' or that annotating any new language will be trivial. But we argue that building upon a highly multilingual syntactic framework (UD) and adopting a fairly neutral representation of meaning (LFs) provides a solid foundation for developing syntactically and semantically rich resources for child-directed speech in new languages, and facilitates cross-linguistic comparison as well.

\section{Conclusion}

Cross-linguistically consistent linguistic annotation of child-directed speech is essential for corpus studies and
computational modeling of child language acquisition. We have presented a methodology for syntactic annotation
on CDS using Universal Dependencies and a conversion method for transducing logical forms from the resulting trees.
We show that the methodology can be reliably applied to English and Hebrew, and propose a way to address common phenomena
in CDS that are scarce in standard UD corpora. 
We then turn to a discussion of the limitations of the current method, suggesting paths for future improvement.
Finally, we apply the proposed methodology to two corpora from CHILDES, the English Adam corpus and the Hebrew Hagar corpus,
yielding sizable, cross-linguistically consistent annotated resources.

While the ability of computational models of acquisition to generalize to different languages is a basic requirement, it has seldom been evaluated empirically, much due to the unavailability of
relevant resources. 
This work immediately enables such comparative investigation in Hebrew and English. Moreover, given the cross-linguistic applicability of UD and the generality of the conversion method, this work is likely to
lead to the compilation of similar resources for many languages more,
thus supporting broadly cross-linguistic corpus research on child-directed speech.
Previous work \citep{Abend:17} showed that a model of a child's acquisition of grammar can be induced from semantic annotation of the kind discussed here. We apply their model to the compiled corpora as a preliminary demonstration of the possibility of comparative computational research on grammar acquisition in the two languages.

 \section*{Declarations}

\paragraph{Conflicts of interest/Competing interests.}
The authors have no conflicts of interest or competing interests that relate to this manuscript.

\paragraph{Graphics program.} All graphs were created using the Python seaborn package (\url{https://seaborn.pydata.org/}).

\section*{Acknowledgements}
 The project SEMANTAX has received funding from the European Research Council (ERC) under the European Union’s Horizon 2020 research and innovation programme (grant agreement No. 742137). The research was funded in part by a James S. McDonnell Foundation Scholar Award (\#220020374) to Sharon Goldwater, and grants by the Israel Science Foundation (grant No. 929/17 and 2424/21) to Omri Abend.
 We would like to acknowledge the contribution of the annotators employed in this project: Zohar Afek, Tamar Johnson, Yelena Lisuk and Adi Shamir, and the contribution of Yael Katz, who partnered in the preprocessing of the Hebrew annotation.

\bibliographystyle{lre_style/spbasic}      
\bibliography{udchildes,sam_bib}   

\begin{thebibliography}{63}
\providecommand{\natexlab}[1]{#1}
\providecommand{\url}[1]{{#1}}
\providecommand{\urlprefix}{URL }
\expandafter\ifx\csname urlstyle\endcsname\relax
  \providecommand{\doi}[1]{DOI~\discretionary{}{}{}#1}\else
  \providecommand{\doi}{DOI~\discretionary{}{}{}\begingroup
  \urlstyle{rm}\Url}\fi
\providecommand{\eprint}[2][]{\url{#2}}

\bibitem[{Abend and Rappoport(2013)}]{abend2013universal}
Abend O, Rappoport A (2013) {U}niversal {C}onceptual {C}ognitive {A}nnotation
  ({UCCA}). In: Proc. of ACL, pp 228--238,
  \urlprefix\url{http://aclweb.org/anthology/P13-1023}

\bibitem[{Abend et~al.(2017)Abend, Kwiatkowski, Smith, Goldwater, and
  Steedman}]{Abend:17}
Abend O, Kwiatkowski T, Smith N, Goldwater S, Steedman M (2017) Bootstrapping
  language acquisition. Cognition 164:116--143

\bibitem[{Alishahi and Stevenson(2008)}]{alishahi2008computational}
Alishahi A, Stevenson S (2008) A computational model of early argument
  structure acquisition. Cognitive science 32(5):789--834

\bibitem[{Banarescu et~al.(2012)Banarescu, Bonial, Cai, Georgescu, Griffitt,
  Hermjakob, Knight, Koehn, Palmer, and Schneider}]{Banarescu:12}
Banarescu L, Bonial C, Cai S, Georgescu M, Griffitt K, Hermjakob U, Knight K,
  Koehn P, Palmer M, Schneider N (2012) Abstract meaning representation (amr)
  1.0 specification

\bibitem[{Banarescu et~al.(2013)Banarescu, Bonial, Cai, Georgescu, Griffitt,
  Hermjakob, Knight, Koehn, Palmer, and Schneider}]{amr}
Banarescu L, Bonial C, Cai S, Georgescu M, Griffitt K, Hermjakob U, Knight K,
  Koehn P, Palmer M, Schneider N (2013) Abstract {M}eaning {R}epresentation for
  sembanking. In: Proc. of the 7th Linguistic Annotation Workshop and
  Interoperability with Discourse, Sofia, Bulgaria, pp 178--186,
  \urlprefix\url{http://www.aclweb.org/anthology/W13-2322}

\bibitem[{Berman(1990)}]{berman1990acquiring}
Berman RA (1990) On acquiring an {(S)VO} language: Subjectless sentences in
  children’s {H}ebrew. Linguistics 28(6):1135--1166

\bibitem[{Berzak et~al.(2016{\natexlab{a}})Berzak, Huang, Barbu, Korhonen, and
  Katz}]{berzak2016anchoring}
Berzak Y, Huang Y, Barbu A, Korhonen A, Katz B (2016{\natexlab{a}}) Anchoring
  and agreement in syntactic annotations. In: Proceedings of the 2016
  Conference on Empirical Methods in Natural Language Processing, Association
  for Computational Linguistics, pp 2215--2224, \doi{10.18653/v1/D16-1239},
  \urlprefix\url{http://aclweb.org/anthology/D16-1239}

\bibitem[{Berzak et~al.(2016{\natexlab{b}})Berzak, Kenney, Spadine, Wang, Lam,
  Mori, Garza, and Katz}]{berzak2016universal}
Berzak Y, Kenney J, Spadine C, Wang JX, Lam L, Mori KS, Garza S, Katz B
  (2016{\natexlab{b}}) Universal {D}ependencies for learner {E}nglish. In:
  Proc. of {ACL}, Berlin, Germany, pp 737--746,
  \urlprefix\url{http://www.aclweb.org/anthology/P16-1070}

\bibitem[{Blodgett and Schneider(2019)}]{blodgett-19}
Blodgett A, Schneider N (2019) An improved approach for semantic graph
  composition with {CCG}. In: Proc. of {IWCS}, Gothenburg, Sweden, pp 55--70,
  \urlprefix\url{https://www.aclweb.org/anthology/W19-0405}

\bibitem[{Blodgett and Schneider(2021)}]{leamr}
Blodgett A, Schneider N (2021) Probabilistic, structure-aware algorithms for
  improved variety, accuracy, and coverage of {AMR} alignments. In: Proc. of
  {ACL-IJCNLP}, Online, pp 3310--3321,
  \urlprefix\url{https://aclanthology.org/2021.acl-long.257}

\bibitem[{Bouma et~al.(2001)Bouma, Noord, and Malouf}]{bouma2001alpino}
Bouma G, Noord GV, Malouf R (2001) Alpino: Wide-coverage computational analysis
  of {D}utch. Language and Computers 37:45--59

\bibitem[{Bowerman(1973)}]{Bowerman:73b}
Bowerman M (1973) Structural relationships in children's utterances: Syntactic
  or semantic? In: Cognitive Development and the Acquisition of Language,
  Academic Press

\bibitem[{Bowerman(1974)}]{Bowerman:74}
Bowerman M (1974) Learning the structure of causative verbs: A study in the
  relationship of cognitive, semantic and syntactic development. Papers and
  Reports on Child Language Development 8:142--178

\bibitem[{Briscoe(2000)}]{Bris:00}
Briscoe T (2000) Grammatical acquisition: Inductive bias and coevolution of
  language and the language acquisition device. Language 76:245--296

\bibitem[{Brown(1973)}]{Brow:73}
Brown R (1973) A First Language: the Early Stages. Harvard University Press,
  Cambridge, MA

\bibitem[{Buttery(2006)}]{Buttery:06}
Buttery P (2006) Computational models for first language acquisition. PhD
  thesis, University of Cambridge

\bibitem[{Culicover and Wilkins(1984)}]{Culi:84}
Culicover P, Wilkins W (1984) Locality in Linguistic Theory. Academic Press,
  New York

\bibitem[{Davidson(1967)}]{Davi:67}
Davidson D (1967) The logical form of action sentences. In: Rescher N (ed) The
  Logic of Decision and Action, University of Pittsburgh Press, Pittsburgh,~PA,
  pp 81--95

\bibitem[{Dirix et~al.(2017)Dirix, Augustinus, van Niekerk, and
  Van~Eynde}]{dirix2017universal}
Dirix P, Augustinus L, van Niekerk D, Van~Eynde F (2017) Universal
  {D}ependencies for {A}frikaans. In: Proceedings of the NoDaLiDa 2017 Workshop
  on Universal Dependencies, 135, pp 38--47

\bibitem[{Fazly et~al.(2010)Fazly, Alishahi, and Stevenson}]{fazl:10}
Fazly A, Alishahi A, Stevenson S (2010) A probabilistic computational model of
  cross-situational word learning. Cognitive Science 34:1017--1063

\bibitem[{Gerdes(2013)}]{Gerdes:13}
Gerdes K (2013) Collaborative dependency annotation. In: Proceedings of the
  Second International Conference on Dependency Linguistics (DepLing 2013), pp
  88--97

\bibitem[{Gretz et~al.(2015)Gretz, Itai, MacWhinney, Nir, and
  Wintner}]{Gretz:15}
Gretz S, Itai A, MacWhinney B, Nir B, Wintner S (2015) Parsing {H}ebrew
  {CHILDES} transcripts. Language Resources and Evaluation 49(1):107--145

\bibitem[{Groschwitz et~al.(2018)Groschwitz, Lindemann, Fowlie, Johnson, and
  Koller}]{groschwitz-18}
Groschwitz J, Lindemann M, Fowlie M, Johnson M, Koller A (2018) {AMR}
  dependency parsing with a typed semantic algebra. In: Proc. of {ACL},
  Melbourne, Australia, pp 1831--1841,
  \urlprefix\url{http://aclweb.org/anthology/P18-1170}

\bibitem[{Gr{\"u}newald et~al.(2021)Gr{\"u}newald, Piccirilli, and
  Friedrich}]{grunewald-etal-2021-coordinate}
Gr{\"u}newald S, Piccirilli P, Friedrich A (2021) Coordinate constructions in
  {E}nglish enhanced {U}niversal {D}ependencies: Analysis and computational
  modeling. In: Proceedings of the 16th Conference of the European Chapter of
  the Association for Computational Linguistics: Main Volume, Association for
  Computational Linguistics, Online, pp 795--809,
  \urlprefix\url{https://aclanthology.org/2021.eacl-main.67}

\bibitem[{Hershcovich et~al.(2019)Hershcovich, Abend, and
  Rappoport}]{hershcovich-etal-2019-content}
Hershcovich D, Abend O, Rappoport A (2019) Content differences in syntactic and
  semantic representation. In: Proceedings of the 2019 Conference of the North
  {A}merican Chapter of the Association for Computational Linguistics: Human
  Language Technologies, Volume 1 (Long and Short Papers), Association for
  Computational Linguistics, Minneapolis, Minnesota, pp 478--488,
  \doi{10.18653/v1/N19-1047},
  \urlprefix\url{https://www.aclweb.org/anthology/N19-1047}

\bibitem[{Hoff-Ginsberg(1985)}]{hoff1985some}
Hoff-Ginsberg E (1985) Some contributions of mothers' speech to their
  children's syntactic growth. Journal of Child Language 12(2):367--385

\bibitem[{Huebner et~al.(2021)Huebner, Sulem, Cynthia, and
  Roth}]{huebner-etal-2021-babyberta}
Huebner PA, Sulem E, Cynthia F, Roth D (2021) {B}aby{BERT}a: Learning more
  grammar with small-scale child-directed language. In: Proceedings of the 25th
  Conference on Computational Natural Language Learning, Association for
  Computational Linguistics, Online, pp 624--646,
  \doi{10.18653/v1/2021.conll-1.49},
  \urlprefix\url{https://aclanthology.org/2021.conll-1.49}

\bibitem[{Kamp and Reyle(1993)}]{Kamp:93}
Kamp H, Reyle U (1993) From Discourse to Logic. Kluwer, Dordrecht

\bibitem[{Kwiatkowski et~al.(2012)Kwiatkowski, Goldwater, Zettlemoyer, and
  Steedman}]{Kwia:12}
Kwiatkowski T, Goldwater S, Zettlemoyer L, Steedman M (2012) A probabilistic
  model of syntactic and semantic acquisition from child-directed utterances
  and their meanings. In: Proceedings of the 13th Conference of the European
  Chapter of the {ACL} ({EACL} 2012), ACL, Avignon, pp 234--244

\bibitem[{Levy and Andrew(2006)}]{levy-06}
Levy R, Andrew G (2006) Tregex and {T}surgeon: tools for querying and
  manipulating tree data structures. In: Proc. of {LREC}, Genoa, Italy, pp
  2231--2234

\bibitem[{Liu and Prud'hommeaux(2021)}]{liu-21}
Liu Z, Prud'hommeaux E (2021) Dependency parsing evaluation for low-resource
  spontaneous speech. In: Proc. of the Second Workshop on Domain Adaptation for
  {NLP}, Kyiv, Ukraine, pp 156--165,
  \urlprefix\url{https://aclanthology.org/2021.adaptnlp-1.16}

\bibitem[{MacWhinney(2000)}]{MacW:00}
MacWhinney B (2000) The {C}{H}{I}{L}{D}{E}{S} Project: Tools for Analyzing
  Talk. Erlbaum, Mahwah,~NJ

\bibitem[{Mao et~al.(2021)Mao, Shi, Wu, Levy, and Tenenbaum}]{Mao:21a}
Mao J, Shi F, Wu J, Levy R, Tenenbaum J (2021) Grammar-based grounded lexicon
  learning. In: Proceedings of the Thirty-Fifth Conference on Neural
  Information Processing Systems, pp 7865--7878

\bibitem[{Marcus et~al.(1993)Marcus, Santorini, and Marcinkiewicz}]{Marc:93}
Marcus M, Santorini B, Marcinkiewicz M (1993) Building a large annotated corpus
  of {E}nglish: The {P}enn {T}reebank. Computational Linguistics 19:313--330

\bibitem[{de~Marneffe et~al.(2021)de~Marneffe, Manning, Nivre, and
  Zeman}]{de_marneffe-21}
de~Marneffe M, Manning CD, Nivre J, Zeman D (2021) Universal {D}ependencies.
  Computational Linguistics 47(2):255--308,
  \urlprefix\url{https://doi.org/10.1162/coli_a_00402}

\bibitem[{{McDonald} et~al.(2013){McDonald}, Nivre, {Quirmbach-Brundage},
  Goldberg, Das, Ganchev, Hall, Petrov, Zhang, T\"{a}ckstr\"{o}m, Bedini,
  Bertomeu~Castell\'{o}, and Lee}]{mcdonald2013universal}
{McDonald} R, Nivre J, {Quirmbach-Brundage} Y, Goldberg Y, Das D, Ganchev K,
  Hall K, Petrov S, Zhang H, T\"{a}ckstr\"{o}m O, Bedini C,
  Bertomeu~Castell\'{o} N, Lee J (2013) Universal dependency annotation for
  multilingual parsing. In: Proc. of {ACL}, Sofia, Bulgaria, pp 92--97,
  \urlprefix\url{http://www.aclweb.org/anthology/P13-2017}

\bibitem[{McNeill(1966)}]{McNeill:66}
McNeill D (1966) Developmental psycholinguistics. In: Smith F, Miller G (eds)
  The Genesis of Language, Cambridge: MIT Press, pp 15--84

\bibitem[{Moon et~al.(2018)Moon, Christodoulopoulos, Fisher, Franco, and
  Roth}]{moon2018gold}
Moon L, Christodoulopoulos C, Fisher C, Franco S, Roth D (2018) Gold standard
  annotations for preposition and verb sense with semantic role labels in
  adult-child interactions. In: Proceedings of the 27th International
  Conference on Computational Linguistics, Association for Computational
  Linguistics, Santa Fe, New Mexico, USA, pp 3004--3014,
  \urlprefix\url{https://www.aclweb.org/anthology/C18-1254}

\bibitem[{Newport(1977)}]{Newp:77}
Newport E (1977) Mother, {I}'d rather do it by myself: Some effects and
  non-effects of maternal speech-style. In: Snow C, Fergusson C (eds) Talking
  to Children:Language Input and Acquisition, Cambridge University Press,
  Cambridge, pp 109--149

\bibitem[{Nguyen(2018)}]{nguyen2018bktreebank}
Nguyen KH (2018) Bktreebank: Building a {V}ietnamese dependency treebank. In:
  LREC, pp 2164--2168

\bibitem[{Nivre et~al.(2016)Nivre, de~Marneffe, Ginter, Goldberg, Hajic,
  Manning, McDonald, Petrov, Pyysalo, Silveira, Tsarfaty, and
  Zeman}]{nivre2016universal}
Nivre J, de~Marneffe MC, Ginter F, Goldberg Y, Hajic J, Manning CD, McDonald R,
  Petrov S, Pyysalo S, Silveira N, Tsarfaty R, Zeman D (2016) Universal
  dependencies v1: A multilingual treebank collection. In: Proc. of LREC, pp
  1659--1666

\bibitem[{Odijk et~al.(2018)Odijk, Dimitriadis, Van~der Klis, Van~Koppen,
  Otten, and Van~der Veen}]{odijk2018anncor}
Odijk J, Dimitriadis A, Van~der Klis M, Van~Koppen M, Otten M, Van~der Veen R
  (2018) The {AnnCor} {CHILDES} treebank. In: Proceedings of LREC

\bibitem[{Pearl and Sprouse(2013)}]{pearl2013syntactic}
Pearl L, Sprouse J (2013) Syntactic islands and learning biases: Combining
  experimental syntax and computational modeling to investigate the language
  acquisition problem. Language Acquisition 20(1):23--68

\bibitem[{Peng and Zeldes(2018)}]{depedit}
Peng S, Zeldes A (2018) All roads lead to {UD}: Converting {S}tanford and
  {P}enn parses to {E}nglish {U}niversal {D}ependencies with multilayer
  annotations. In: Proceedings of the Joint Workshop on Linguistic Annotation,
  Multiword Expressions and Constructions ({LAW}-{MWE}-{C}x{G}-2018), Santa Fe,
  NM, pp 167--177, \urlprefix\url{https://www.aclweb.org/anthology/W18-4918}

\bibitem[{Perfors et~al.(2011)Perfors, Tenenbaum, and Regier}]{Perfors:11b}
Perfors A, Tenenbaum J, Regier T (2011) The learnability of abstract syntactic
  principles. Cognition 118:306--338

\bibitem[{Pinker(1979)}]{Pink:79}
Pinker S (1979) Formal models of language learning. Cognition 7:217--283

\bibitem[{Przepi{\'o}rkowski and
  Patejuk(2019)}]{przepiorkowski-patejuk-2019-nested}
Przepi{\'o}rkowski A, Patejuk A (2019) Nested coordination in {U}niversal
  {D}ependencies. In: Proceedings of the Third Workshop on Universal
  Dependencies (UDW, SyntaxFest 2019), Association for Computational
  Linguistics, Paris, France, pp 58--69, \doi{10.18653/v1/W19-8007},
  \urlprefix\url{https://aclanthology.org/W19-8007}

\bibitem[{Pullum(1990)}]{pullum1990constraints}
Pullum GK (1990) Constraints on intransitive quasi-serial verb constructions in
  modem colloquial {E}nglish. Working Papers in Linguistics 39:218--39

\bibitem[{Pustejovsky(1998)}]{pustejovsky1998generative}
Pustejovsky J (1998) The generative lexicon. MIT press

\bibitem[{Reddy et~al.(2016)Reddy, T{\"a}ckstr{\"o}m, Collins, Kwiatkowski,
  Das, Steedman, and Lapata}]{reddy2016transforming}
Reddy S, T{\"a}ckstr{\"o}m O, Collins M, Kwiatkowski T, Das D, Steedman M,
  Lapata M (2016) Transforming dependency structures to logical forms for
  semantic parsing. Transactions of the Association for Computational
  Linguistics 4:127--140

\bibitem[{Sagae et~al.(2010)Sagae, Davis, Lavie, MacWhinney, and
  Wintner}]{Saga:10}
Sagae K, Davis E, Lavie A, MacWhinney B, Wintner S (2010) Morphosyntactic
  annotation of {CHILDES} transcripts. Journal of Child Language 37:705--729

\bibitem[{Sanguinetti et~al.(2020)Sanguinetti, Cassidy, Bosco,
  \c{C}etino\u{g}lu, Cignarella, Lynn, Rehbein, Ruppenhofer, Seddah, and
  Zeldes}]{sanguinetti-20}
Sanguinetti M, Cassidy L, Bosco C, \c{C}etino\u{g}lu O, Cignarella AT, Lynn T,
  Rehbein I, Ruppenhofer J, Seddah D, Zeldes A (2020) Treebanking
  user-generated content: a {UD} based overview of guidelines, corpora and
  unified recommendations. {arXiv}:201102063 [cs]
  \urlprefix\url{http://arxiv.org/abs/2011.02063}, {arXiv}: 2011.02063

\bibitem[{Savary et~al.(2023)Savary, Stymne, Barbu~Mititelu, Schneider,
  Ramisch, and Nivre}]{parseme-ud-23}
Savary A, Stymne S, Barbu~Mititelu V, Schneider N, Ramisch C, Nivre J (2023)
  {PARSEME} meets {U}niversal {D}ependencies: Getting on the same page in
  representing multiword expressions. Northern European Journal of Language
  Technology 9(1), \doi{10.3384/nejlt.2000-1533.2023.4453},
  \urlprefix\url{https://nejlt.ep.liu.se/article/view/4453}

\bibitem[{Schuster and Manning(2016)}]{Schuster2016enhanced}
Schuster S, Manning CD (2016) Enhanced {E}nglish {U}niversal {D}ependencies: An
  improved representation for natural language understanding tasks. In: Proc.
  of LREC, ELRA,
  \urlprefix\url{https://nlp.stanford.edu/pubs/schuster2016enhanced.pdf}

\bibitem[{Silveira et~al.(2014)Silveira, Dozat, Marneffe, Bowman, Connor,
  Bauer, and Manning}]{silveira2014gold}
Silveira N, Dozat T, Marneffe MD, Bowman SR, Connor M, Bauer J, Manning CD
  (2014) A gold standard dependency corpus for {E}nglish. In: Calzolari N,
  Choukri K, Declerck T, Loftsson H, Maegaard B, Mariani J, Moreno A, Odijk J,
  Piperidis S (eds) Proc. of {LREC}, Reykjav\'{i}k, Iceland, pp 2897--2904,
  \urlprefix\url{http://www.lrec-conf.org/proceedings/lrec2014/pdf/1089_Paper.pdf}

\bibitem[{Steedman(2000{\natexlab{a}})}]{stee:99}
Steedman M (2000{\natexlab{a}}) The Syntactic Process. MIT Press, Cambridge, MA

\bibitem[{Steedman(2000{\natexlab{b}})}]{ccg}
Steedman M (2000{\natexlab{b}}) The {S}yntatic {P}rocess. {MIT} Press,
  Cambridge, {MA}

\bibitem[{Szubert et~al.(2018)Szubert, Lopez, and
  Schneider}]{szubert-etal-2018-structured}
Szubert I, Lopez A, Schneider N (2018) A structured syntax-semantics interface
  for {E}nglish-{AMR} alignment. In: Proceedings of the 2018 Conference of the
  North {A}merican Chapter of the Association for Computational Linguistics:
  Human Language Technologies, Volume 1 (Long Papers), Association for
  Computational Linguistics, New Orleans, Louisiana, pp 1169--1180,
  \doi{10.18653/v1/N18-1106}, \urlprefix\url{https://aclanthology.org/N18-1106}

\bibitem[{Tsarfaty(2013)}]{tsarfaty2013unified}
Tsarfaty R (2013) A unified morpho-syntactic scheme of {S}tanford dependencies.
  In: Proc. of {ACL}, Sofia, Bulgaria, pp 578--584,
  \urlprefix\url{https://www.aclweb.org/anthology/P13-2103}

\bibitem[{Van~Gysel et~al.(2021)Van~Gysel, Vigus, Chun, Lai, Moeller, Yao,
  O'Gorman, Cowell, Croft, Huang, Haji\v{c}, Martin, Oepen, Palmer,
  Pustejovsky, Vallejos, and Xue}]{umr}
Van~Gysel JEL, Vigus M, Chun J, Lai K, Moeller S, Yao J, O'Gorman T, Cowell A,
  Croft W, Huang C, Haji\v{c} J, Martin JH, Oepen S, Palmer M, Pustejovsky J,
  Vallejos R, Xue N (2021) Designing a {U}niform {M}eaning {R}epresentation for
  natural {L}anguage processing. {KI} - K\"{u}nstliche Intelligenz
  35(3):343--360, \urlprefix\url{https://doi.org/10.1007/s13218-021-00722-w}

\bibitem[{Villavicencio(2002)}]{Vill:02}
Villavicencio A (2002) The acquisition of a unification-based generalised
  categorial grammar. PhD thesis, University of Cambridge

\bibitem[{Yang(2002)}]{Yang:02}
Yang C (2002) Knowledge and Learning in Natural Language. Oxford University
  Press, Oxford

\bibitem[{Yedetore et~al.(2023)Yedetore, Linzen, Frank, and
  McCoy}]{aditya2023how}
Yedetore A, Linzen T, Frank R, McCoy RT (2023) How poor is the stimulus?
  evaluating hierarchical generalization in neural networks trained on
  child-directed speech. \doi{10.48550/ARXIV.2301.11462},
  \urlprefix\url{https://arxiv.org/abs/2301.11462}

\end{thebibliography}

 \section*{Appendix: Hebrew Transcription Conventions}\label{app:heb}

We adopt the  transcription conventions used in the Hagar corpus.
The consonants and their transliterations are:

\begin{table}[H]
\begin{tabular}{ c c }
    \cjRL{'} & \textglotstop{} \\
    \cjRL{b}   & b/v \\
    \cjRL{g} & g \\
    \cjRL{d} & d \\
    \cjRL{h} & h \\
    \cjRL{w} & w \\
    \cjRL{z} & z \\
    \cjRL{.h} & \d{k} \\
    \cjRL{.t} & \d{t} \\
    \cjRL{y} & y \\
    \cjRL{k} & k/\d{k} \\
    \cjRL{l} & l \\
    \cjRL{m} & m \\
    \cjRL{n} & n \\
    \cjRL{s} & s \\
    \cjRL{`} & \textrevglotstop{} \\
    \cjRL{p} & p/f \\
    \cjRL{.s} & c \\
    \cjRL{q} & q \\
    \cjRL{r} & r \\
    \cjRL{+s} & \v{s} \\
    \cjRL{,s}& \d{s} \\
    \cjRL{t} & t \\

\end{tabular}
\end{table}

The vowels in use are (stressed and unstressed): 

\begin{table}[H]
\begin{tabular}{ c c c c c }
\={a} & \={e} & \={\i} & \={o} & \={u} \\
a & e & i & o & u \\
        
\end{tabular}
\end{table}

 \end{document}


\appendix
\section{Supplementary material}

\subsection{Longitudinal analysis}
For most dependencies we observe no regular change in frequency as the child gets older. The constructions for which a trend was detected are discussed in the main paper. Here we present the plots of frequency over time for all dependencies in both corpora. Each graph combines data from the Adam and Hagar corpora to allow for an easier comparison across languages and ages. Notice that the Hagar corpus starts at an earlier age and that the sessions were much more frequent than in the Adam corpus, leading to more jagged plots.

\begin{figure}[H]
\centering
\begin{subfigure}{\textwidth}
  \centering
  \includegraphics[width=\linewidth]{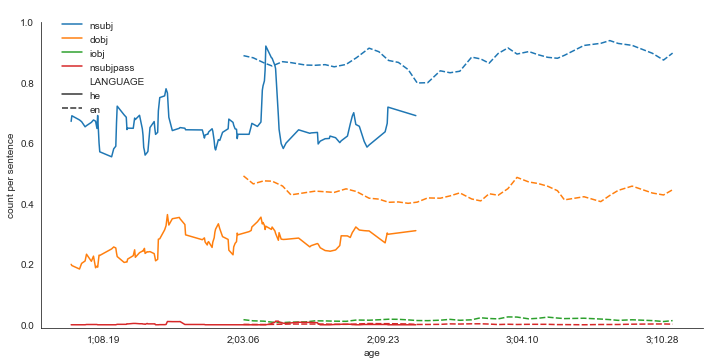}
  \caption{Core arguments}
  \label{fig:args}
\end{subfigure}

\begin{subfigure}{\textwidth}
  \centering
  \includegraphics[width=\linewidth]{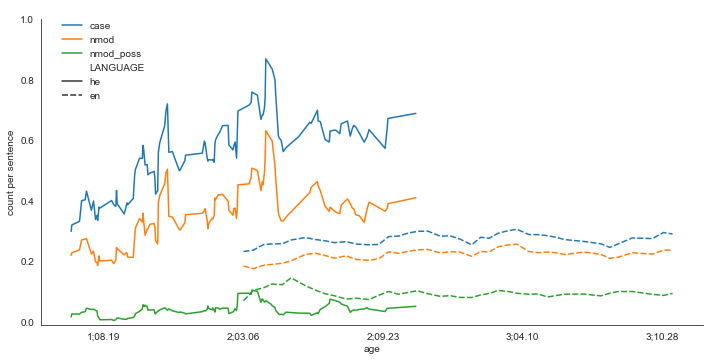}
  \caption{Nominal modifiers, part 1}
  \label{fig:nmod1}
\end{subfigure}

\begin{subfigure}{\textwidth}
  \centering
  \includegraphics[width=\linewidth]{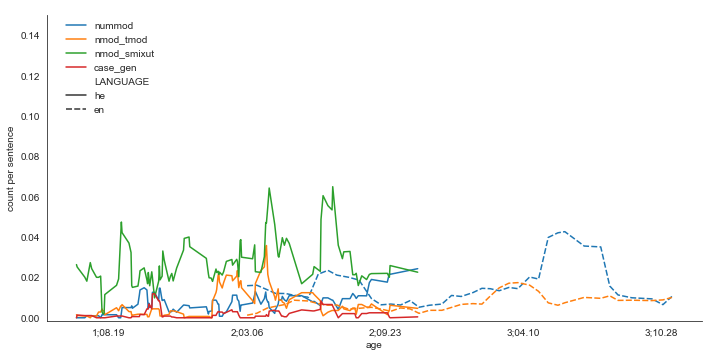}
  \caption{Nominal modifiers, part 2}
  \label{fig:nmod2}
\end{subfigure}

\caption{Proportion of sentences containing given dependency per session in the Adam (dashed) and Hagar (solid) corpora; frequencies are smoothed over 5 sessions.}
\label{fig:longitudinal_1}
\end{figure}

\begin{figure}[H]
\centering
\begin{subfigure}{\textwidth}
  \centering
  \includegraphics[width=\linewidth]{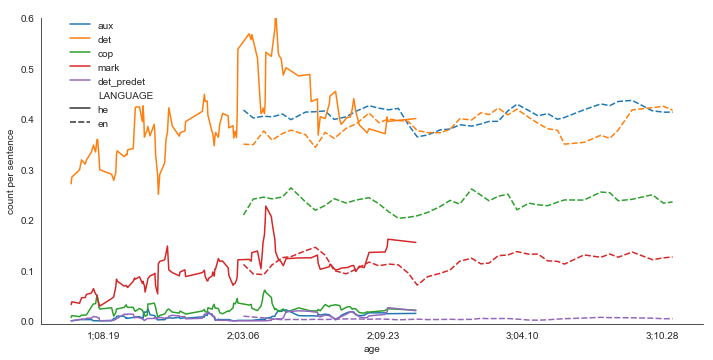}
  \caption{Determiners and purely syntactic relations}
  \label{fig:syntax}
\end{subfigure}

\begin{subfigure}{\textwidth}
  \centering
  \includegraphics[width=\linewidth]{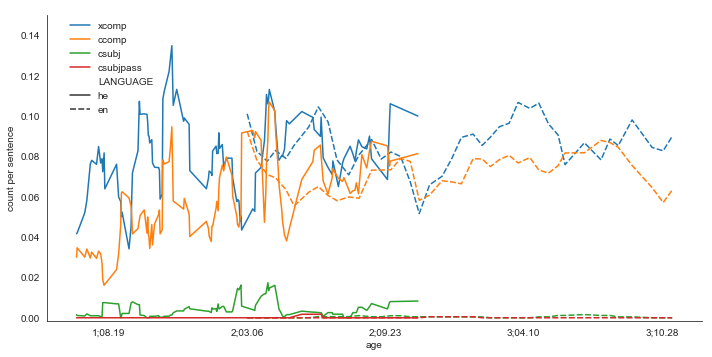}
  \caption{Clausal arguments}
  \label{fig:clausal_arg}
\end{subfigure}

\begin{subfigure}{\textwidth}
  \centering
  \includegraphics[width=\linewidth]{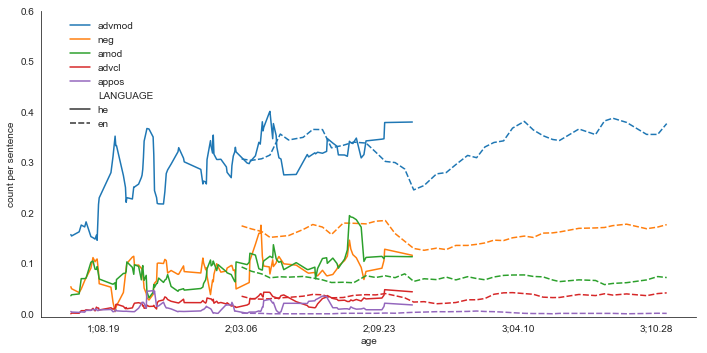}
  \caption{Modifiers of nominals}
  \label{fig:nom_mod}
\end{subfigure}
\caption{Proportion of sentences containing given dependency per session in the Adam (dashed) and Hagar (solid) corpora; frequencies are smoothed over 5 sessions.}
\label{fig:longitudinal_2}
\end{figure}

\begin{figure}[H]
\centering
\begin{subfigure}{\textwidth}
  \centering
  \includegraphics[width=\linewidth]{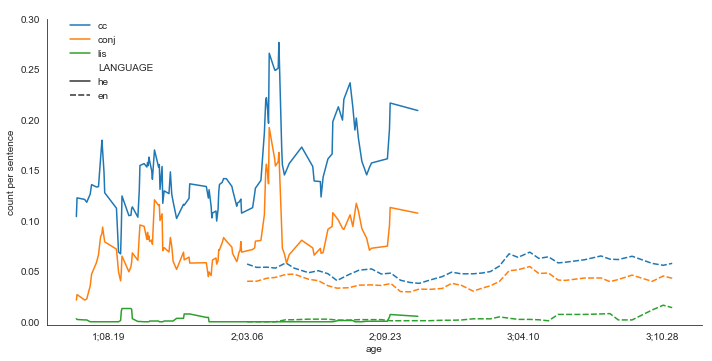}
  \caption{Conjunction}
  \label{fig:conj}
\end{subfigure}

\begin{subfigure}{\textwidth}
  \centering
  \includegraphics[width=\linewidth]{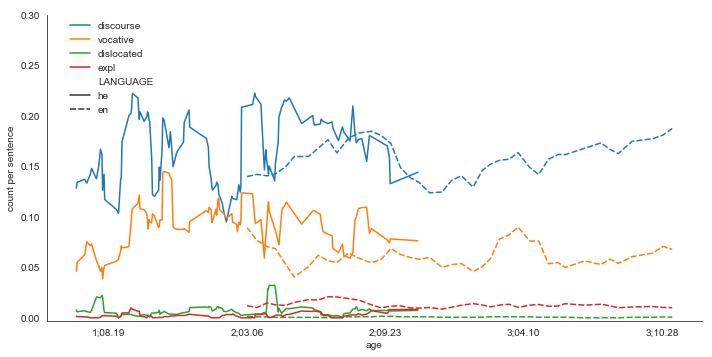}
  \caption{Discursive relations, part 1}
  \label{fig:discourse_1}
\end{subfigure}

\begin{subfigure}{\textwidth}
  \centering
  \includegraphics[width=\linewidth]{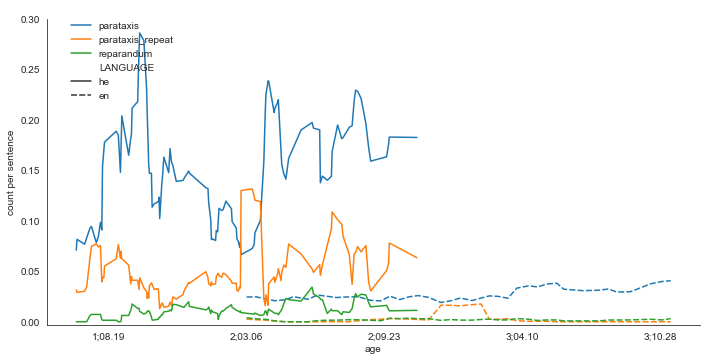}
  \caption{Discursive relations, part 2}
  \label{fig:discourse_2}
\end{subfigure}
\caption{Proportion of sentences containing given dependency per session in the Adam (dashed) and Hagar (solid) corpora; frequencies are smoothed over 5 sessions.}
\label{fig:longitudinal_3}
\end{figure}

\begin{figure}[H]
\centering
\begin{subfigure}{\textwidth}
  \centering
  \includegraphics[width=\linewidth]{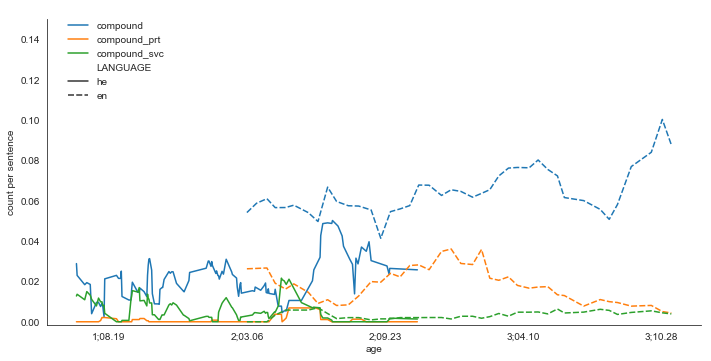}
  \caption{Compounding, part 1}
  \label{fig:compound_1}
\end{subfigure}

\begin{subfigure}{\textwidth}
  \centering
  \includegraphics[width=\linewidth]{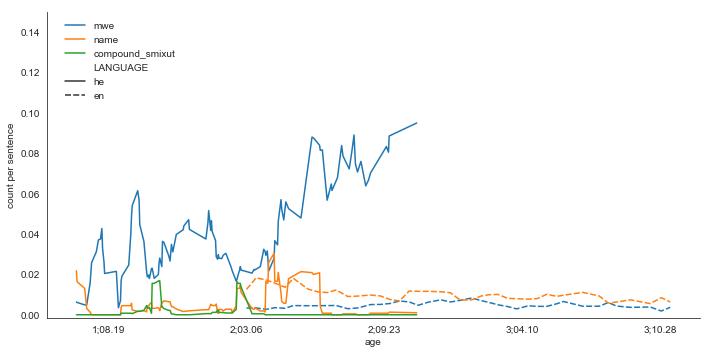}
  \caption{Compounding, part 2}
  \label{fig:compound_2}
\end{subfigure}

\begin{subfigure}{\textwidth}
  \centering
  \includegraphics[width=\linewidth]{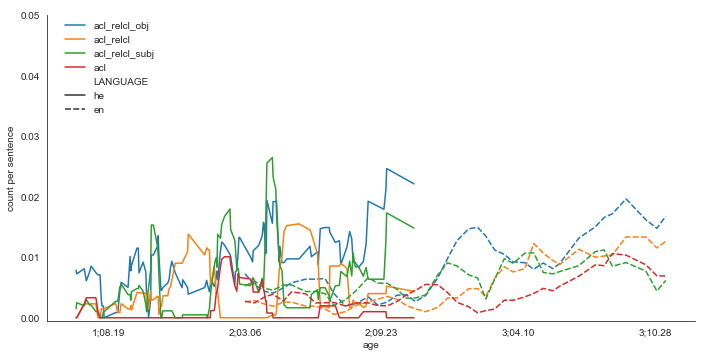}
  \caption{Relative clause}
  \label{fig:relcl}
\end{subfigure}
\caption{Proportion of sentences containing given dependency per session in the Adam (dashed) and Hagar (solid) corpora; frequencies are smoothed over 5 sessions.}
\label{fig:longitudinal_4}
\end{figure}

\begin{figure}[H]
\centering
\begin{subfigure}{\textwidth}
  \centering
  \includegraphics[width=\linewidth]{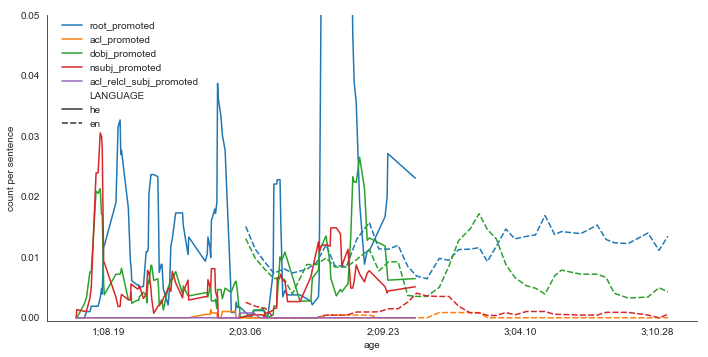}
  \caption{Ellipsis, part 1}
  \label{fig:ellipsis_1}
\end{subfigure}

\begin{subfigure}{\textwidth}
  \centering
  \includegraphics[width=\linewidth]{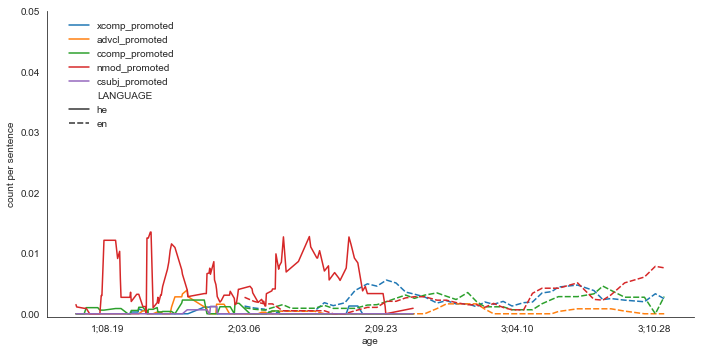}
  \caption{Ellipsis, part 2}
  \label{fig:ellipsis_2}
\end{subfigure}
\caption{Proportion of sentences containing given dependency per session in the Adam (dashed) and Hagar (solid) corpora; frequencies are smoothed over 5 sessions.}
\label{fig:longitudinal_5}
\end{figure}